\documentclass[journal]{IEEEtran}
\usepackage{xcolor,soul,framed} 

\colorlet{shadecolor}{yellow}
\usepackage[pdftex]{graphicx}
\graphicspath{{../pdf/}{../jpeg/}{../svg/}}
\DeclareGraphicsExtensions{.pdf,.jpeg,.png,.svg}

\usepackage[cmex10]{amsmath}
\usepackage{array}
\usepackage{mdwmath}
\usepackage{mdwtab}
\usepackage{eqparbox}
\usepackage{url}
\usepackage{float}
\usepackage{amssymb}
\usepackage{makecell}
\usepackage{cite}
\usepackage{tabularx}
\usepackage{amsfonts}
\usepackage{multicol}
\usepackage{hyperref}

\usepackage{stfloats}

\begin{document}
\bstctlcite{IEEEexample:BSTcontrol}
    \title{ Wavelet Transform-assisted Adaptive Generative Modeling for Colorization}
  \author{Jin~Li, Wanyun~Li, Zichen~Xu, Yuhao~Wang,~\IEEEmembership{Senior Member,~IEEE,}
      \\Qiegen~Liu,~\IEEEmembership{Senior Member,~IEEE}
\vspace{-0.5cm}

  \thanks{This work was supported in part by the National Natural Science Foundation of China under 61871206. J. Li and W. Li are co-first authors. (\emph{Corresponding authors: Yuhao Wang, Qiegen Liu.)}}
  \thanks{J. Li, W. Li, Z. Xu, Y. Wang and Q. Liu are with the Information Engineering School, Nanchang University, Nanchang 330031, China. {(\{lijin, liwanyun\}@email.ncu.edu.cn, \{xuz, wangyuhao, liuqiegen\}@ncu.edu.cn)}}
 }

\markboth{
}{}
\maketitle

\begin{abstract}
Unsupervised deep learning has recently demonstrated the promise of producing high-quality samples. While it has tremendous potential to promote the image colorization task, the performance is limited owing to the high-dimension of data manifold and model capability. This study presents a novel scheme that exploits the score-based generative model in wavelet domain to address the issues. By taking advantage of the multi-scale and multi-channel representation via wavelet transform, the proposed model learns the richer priors from stacked coarse and detailed wavelet coefficient components jointly and effectively. This strategy also reduces the dimension of the original manifold and alleviates the curse of dimensionality, which is beneficial for estimation and sampling. Moreover, dual consistency terms in the wavelet domain, namely data-consistency and structure-consistency are devised to leverage colorization task better. Specifically, in the training phase, a set of multi-channel tensors consisting of wavelet coefficients is used as the input to train the network with denoising score matching. In the inference phase, samples are iteratively generated via annealed Langevin dynamics with data and structure consistencies. Experiments demonstrated remarkable improvements of the proposed method on both generation and colorization quality, particularly in colorization robustness and diversity.
\end{abstract}

\begin{IEEEkeywords}
Automatic colorization, Wavelet transform, Unsupervised learning, Generative model, Multi-scale.
\end{IEEEkeywords}

%
\IEEEpeerreviewmaketitle


\section{Introduction}

\IEEEPARstart{I}{MAGE} colorization, the process of adding color to an original grayscale image, has many practical applications in the computer vision and graphics community \cite{fatima2021grey,baig2017multiple,bian2021deep}. As the colorization problem requires a mapping from a one-channel grayscale image to a multi-channel composite image, it is essentially ill-conditioned and ambiguous with multi-modal uncertainty. 

Over the past decades, many approaches, including earlier attempts that required user interaction (e.g., scribble-based\cite{levin2004colorization,huang2005adaptive,qu2006manga,luan2007natural} or example-based methods\cite{welsh2002transferring,ironi2005colorization,charpiat2008automatic,chia2011semantic}) and automatic learning-based methods\cite{deshpande2015learning,yoo2019coloring,suarez2017infrared,vitoria2020chromagan,zhou2020progressive,cao2017unsupervised,zhang2016colorful,iizuka2016let,isola2017image,zhao2020pixelated}, have been developed to tackle the issue of colorization. Among them, traditional methods rely on significant user effort and time to achieve proper results. The supervised methods have disadvantages of demanding a large quantity of labeled training datasets and producing monotonous colorization results. Therefore, some unsupervised learning techniques have been heavily investigated in these years. The most prevailing methods use generative adversarial network (GAN) or variational auto-encoder (VAE). For instance, Yoo \emph{et al.} \cite{yoo2019coloring} proposed a model called MemoPainter that can produce high-quality colorization with limited data via GAN and memory networks. Suarez \emph{et al.} \cite{suarez2017infrared} used a triplet model based on GAN architecture for learning each of color channels independently in a more homogeneous way. Deshpande \emph{et al.}\cite{deshpande2017learning} employed VAE to yield multiple diverse yet realistic colorization. Recently, some underlying theoretic schemes concerning denoising score matching (DSM) \cite{vincent2008extracting,vincent2011connection} were reported by different research groups. Jayaram \emph{et al.}\cite{jayaram2020source} made a preliminary attempt that treating the colorization task as a color channel separation problem, and proposed a “BASIS” separation method based on Noise Conditional Score Networks (NCSN)\cite{song2019generative} using DSM. NCSN is a kind of explicit generative model where samples are produced progressively via Langevin dynamics using score—the gradient of logarithm of probability density which is estimated by DSM. Remarkably, NCSN can estimate and sample explicitly without adversarial optimization, and can produce realistic images that rival GANs.

\begin{figure*}
\setlength{\belowcaptionskip}{-3pt}
  \begin{center}
  \includegraphics[width=7in]{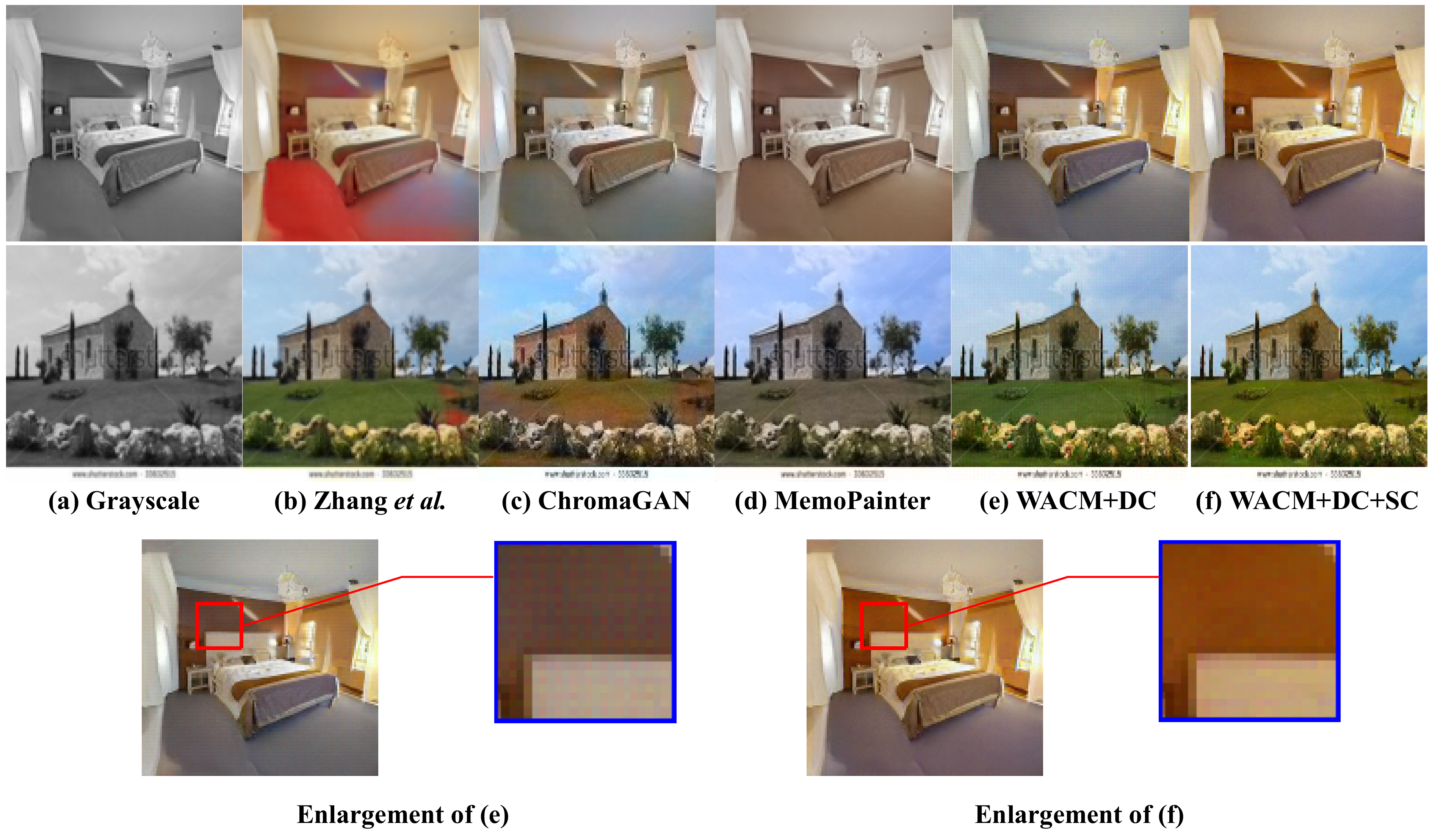}\\
  \vspace{-0.4cm}
  \caption{\hspace{-1.2em} Visual comparison of Zhang \emph{et al.} (b), ChromaGAN (c), MemoPainter (d) and WACM with different constraints (e, f). It can be observed in the first line that Zhang \emph{et al.} (b) assigns unreasonable colors to objects, such as the red floor and blue wall. Meanwhile, the results of ChromaGAN (c) and MemoPainter (d) in the second line suffer from color pollution and desaturation respectively. In this work, dual consistency terms are introduced to leverage the generative model in wavelet domain. The Data-Consistency (DC) in (e) is enforced to achieve a basic proper colorization. Additionally, by enforcing DC and Structure-Consistency (SC) simultaneously, some gridding effects shown in (e) can be eliminated to achieve a better colorization image as in (f). The proposed WACM model involved with two consistencies can attain a high-quality colorization performance.}\label{fimg}
  \end{center}
\setlength{\belowcaptionskip}{-1pt}
\vspace{-0.8cm}
\end{figure*}

Currently, the major deficiencies of score matching based generative models include the short of model capability and the curse of manifold dimensionality \cite{narayanan2010sample,rifai2011manifold}. Specifically, both the score estimation of DSM and the sampling speed and quaility of Langevin dynamics are highly correlated with the intrinsic dimension of data manifold. It is difficult for DSM to provide accurate score estimation in high-dimensional pixel space, where the sampling of Langevin dynamics has serious obstacles in mixing time and convergence as well. In fact, a lot of progress has been made in improving the naïve NCSN. Quan \emph{et al.} \cite{quan2021homotopic} employed the channel-copy technique to form an embedded multi-channel tensor to enhance score estimation accuracy. Zhou \emph{et al.} \cite{zhou2021learning} learned a high-dimensional distribution with score estimation under latent neural Fokker-Planck kernels. Notably, Block \emph{et al.} \cite{block2020fast} proposed a multi-resolution strategy based on upsampling to reduce the dimension of data to improve the generative speed and quality. However, their strategy is still enforced in intensity domain and lacks significant progress. Motivated by this, we focus on a more neat and efficient method to construct theoretically lower-dimensional manifold of the images features distribution, thus to improve the performance of generative model. Specifically, we leverage the image generation capability by embedding with specific wavelet kernel, additionally with constraints in the latent space. 

In this work, we aim to combine the score-based generative model with the Discrete Wavelet Transform (DWT) for generation and colorization task. DWT \cite{akansu2001multiresolution,zhang2019wavelet,acharya2020image,guo2017deep,sharma2016satellite,ghazali2007feature,zhu1998study,chowdhury2012image} is a well-known tool in image processing, which allows images to be decomposed into elementary form at different positions and scales, and subsequently reconstructed with high precision. It has been widely applied to various image processing tasks. For example, Acharya \emph{et al.}\cite{acharya2020image} proposed an image classification method that processes the input with DWT, which can reduce the analyzing time and increase the accuracy. Guo \emph{et al.}\cite{guo2017deep} suggested training network in wavelet domain to address image super-resolution problem as well. 

There are several key advantages for introducing DWT into score-based generative model: First, DWT is a powerful mathematical tool for image processing, which provides an efficient characterization of the coarse and detailed frequency spectrums in images. The richer, structured statistical information of an image contained in wavelet domain is beneficial for the model to learn prior information than in intensity domain. Second, DWT provides a multi-scale downsampling representation of an image, which effectively reduces the inherent dimension of the data manifold, especially for high-resolution images with complex patterns. In addition, Liu \emph{et al.} \cite{liu2020highly} proposed that the inputs of Denoising AutoEncoder (DAE) with more channels  contribute to the network learning capability and the subsequent recovery ability, which is also consistent with the phenomenon that the prior information learned from multi-channel images is more effective than that from the single‐channel counterpart in image restoration tasks. On this basis, the separate wavelet coefficients are processed into a multi-channel feature. 
To sum up, by exploiting the multi-scale and multi-channel feature aggregation via wavelet transform, the proposed strategy provides a lower-dimensional but more informative representation for generative modeling in wavelet domain, which greatly facilitates both the score estimation and Langevin dynamics, and finally improves model capability and generation performance.

\begin{table*}[ht]
\small
\renewcommand{\arraystretch}{1.2}
    \centering
    \caption{ Summary of related colorization methods.}
    \vspace{-0.2cm}
    \begin{tabular}{c|c|c|c|c}
    \hline \hline
        \textbf{Method} & \textbf{ Model} & \textbf{Diversity} & \textbf{Description} & \textbf{Disadvantages} \\
\hline\hline

Zhang \emph{et al.}\cite{zhang2016colorful} & CNN & ${\times}$& A classification network with re-balancing &  Speckle noise \\\hline

Iizuka \emph{et al.}\cite{iizuka2016let} & CNN &${\times}$ & Has two branches to learn features at multiple scales
 & Produce single color \\
 
\hline

Isola \emph{et al.}\cite{isola2017image} & GAN&${\times}$& A GAN conditional network with L1 loss
 & “Mode collapse” problem\\\hline

Cao \emph{et al.}\cite{cao2017unsupervised} &GAN&${\times}$& Adding noise channels to solving “mode collapse”
 & Exist colored noise\\\hline
 
Memopainter\cite{yoo2019coloring} & GAN & ${\checkmark}$ & Integrating memory network with networks & Low saturation outputs\\\hline

ChromaGAN\cite{vitoria2020chromagan} & GAN& ${\checkmark}$& Incorporating perceptual and semantic features & Color bleeding \\

\hline
Deshpande \emph{et al.}\cite{deshpande2017learning} & VAE & \checkmark & Using mixture density network and VAE’s decoder & Blurry and sepia toned outputs \\\hline

iGM-6C\cite{zhou2020progressive} & DSM & \checkmark & Exploring multi-color space prior &  Blurry and low saturation\\
\hline\hline
    \end{tabular}
    \label{tab:my_label}
    \vspace{-0.5cm}
\end{table*}

However, it is necessary to impose some guidelines and consistent constraints to further exploit the generative modeling in wavelet domain for colorization task. Therefore, Data-Consistency (DC) and Structure-Consistency (SC) are devised in this study to solve these issues effectively. Among them, DC can guarantee the basic effect of colorization on the input grayscale images. SC is used to avoid improper effects and improve the colorization performance by ensuring that the generated results in wavelet domain satisfy the strict transformation relationship between the wavelet transform and its inverse process. For example, in Fig. \ref{fimg}(e), we can observe the deficiency of the “gridding” effect that appeared in the colorization results. Benefiting from the multi-scale and multi-channel representation in wavelet domain as well as iteratively generating under the dual consistencies, the proposed Wavelet transform-assisted Adaptive Colorization Model(WACM) performs excellently in various image colorization tasks. In summary, the main contributions of this work are as follows:

\subsubsection{\textbf{New Strategy Design}} It is the first pilot method that exploring score-based generative modeling in wavelet domain. To alleviate the issues of model capability and dimension of manifold, a novel scheme is proposed to exploit the advantages of the wavelet transform. Both the estimation and subsequent sampling are performed in a multi-scale and multi-channel space with lower-dimension, which paves the way for attaining more diverse and high-quality generation and colorization.

\subsubsection{\textbf{Two Efficient Consistencies}} Two consistencies, namely data-consistency and structure-consistency, are devised to facilitate the colorization model in wavelet domain. DC guarantees the basic color performance of the model, and SC helps to achieve better performance. The dual consistencies further improve the adaptability and robustness of colorization.

\subsubsection{\textbf{Remarkable Performance and Diversity}}Combining the above advantages, the proposed WACM achieves highly competitive colorization performance compared to the state-of-the-arts. Sufficient experimental results demonstrate the superiority of WACM in accuracy, naturalness and diversity on multiple benchmark datasets.

Section II provides a brief overview of some relevant works on colorization, 2D-DWT,  DSM and Langevin dynamics. In section III, we elaborate on the formulation of the proposed method and the dual consistencies. Section IV presents the colorization performance of the present model, including comparisons with the state-of-the-arts, ablation study, robustness, and diversity tests, as well as discusses the effects of pre-processing and post-processing on the colorization results and two existing limitations in real-world applications. Conclusion and future works are given in Section V.

\section{Related Work}

\subsection {Image Colorization Techniques}

Image colorization refers to estimating the color information from a grayscale image, which provides a practical solution to enhance old pictures as well as express artistic creativity. In the past two decades, several colorization techniques have been proposed, ranging from user-guided methods \cite{levin2004colorization,huang2005adaptive,qu2006manga,luan2007natural,welsh2002transferring,ironi2005colorization,charpiat2008automatic,chia2011semantic} to automatic learning-based methods\cite{deshpande2015learning,yoo2019coloring,suarez2017infrared,vitoria2020chromagan,zhou2020progressive,cao2017unsupervised,zhang2016colorful,iizuka2016let,isola2017image,deshpande2017learning,zhao2020pixelated}. 

Because colorization is ill-posed and inherently ambiguous, early attempts highly rely on additional user interventions. Considering the amount of user involvement in problem-solving and the way of retrieving the data required, these methods can be roughly categorized into scribble-based\cite{levin2004colorization,huang2005adaptive,qu2006manga,luan2007natural} and example-based\cite{welsh2002transferring,ironi2005colorization,charpiat2008automatic,chia2011semantic}. Scribble-based methods generally formulate colorization as a constrained optimization problem that propagates user-specified color scribbles based on some low-level similarity metrics. Example-based methods focus on colorizing the input grayscale image with the color statistics transferred from a reference. 

Recently, learning-based approaches have demonstrated their effectiveness in image colorization task. Zhang \emph{et al.}\cite{zhang2016colorful} considered colorization as a classification task and proposed a network trained with a multinomial cross entropy loss with class-rebalancing techniques to predict “ab” pairs in Lab color space. Iizuka \emph{et al.} \cite{iizuka2016let} proposed a deep network with a fusion layer that merges local information dependent on small image patches with global priors computed from the entire image.

Due to the diversity of results and the less reliance on structured datasets, unsupervised learning is considered a promising future direction for image colorization \cite{anwar2020image}. Cao \emph{et al.} \cite{cao2017unsupervised} proposed the utilization of conditional GANs for the diverse colorization of real-world objects. They employed five fully convolutional layers with batch normalization and ReLU in the generator of GAN network. However, there is still noise in their results as the method \cite{isola2017image} . Yoo \emph{et al.}\cite{yoo2019coloring} proposed a memory-augmented model MemoPainter that consists of memory networks and colorization networks to produce colorization with limited data. Zhou \emph{et al.} \cite{zhou2020progressive} proposed an iterative generative model which is exploited in multi-color spaces jointly and is enforced with linearly autocorrelative constraint. Victoria \emph{et al.}\cite{vitoria2020chromagan}  exploited features via an end-to-end self-supervised generative adversarial network that learns to colorize by incorporating perceptual and semantic understanding.

\subsection {2D-DWT}
DWT is a well-known tool in image processing community, which is capable of effectively analyzing image features, particularly image details\cite{zhang2019wavelet}. Although wavelets have been applied in a variety of applications such as removing speckle noise from images\cite{sharma2016satellite}, image classification \cite{ghazali2007feature}, texture analysis \cite{ghazali2007feature,zhu1998study} and image compression \cite{ chowdhury2012image}, it has rarely been applied in image colorization.

\begin{figure}
\setlength{\belowcaptionskip}{-3pt}
  \begin{center}
  \includegraphics[width=3.5in]{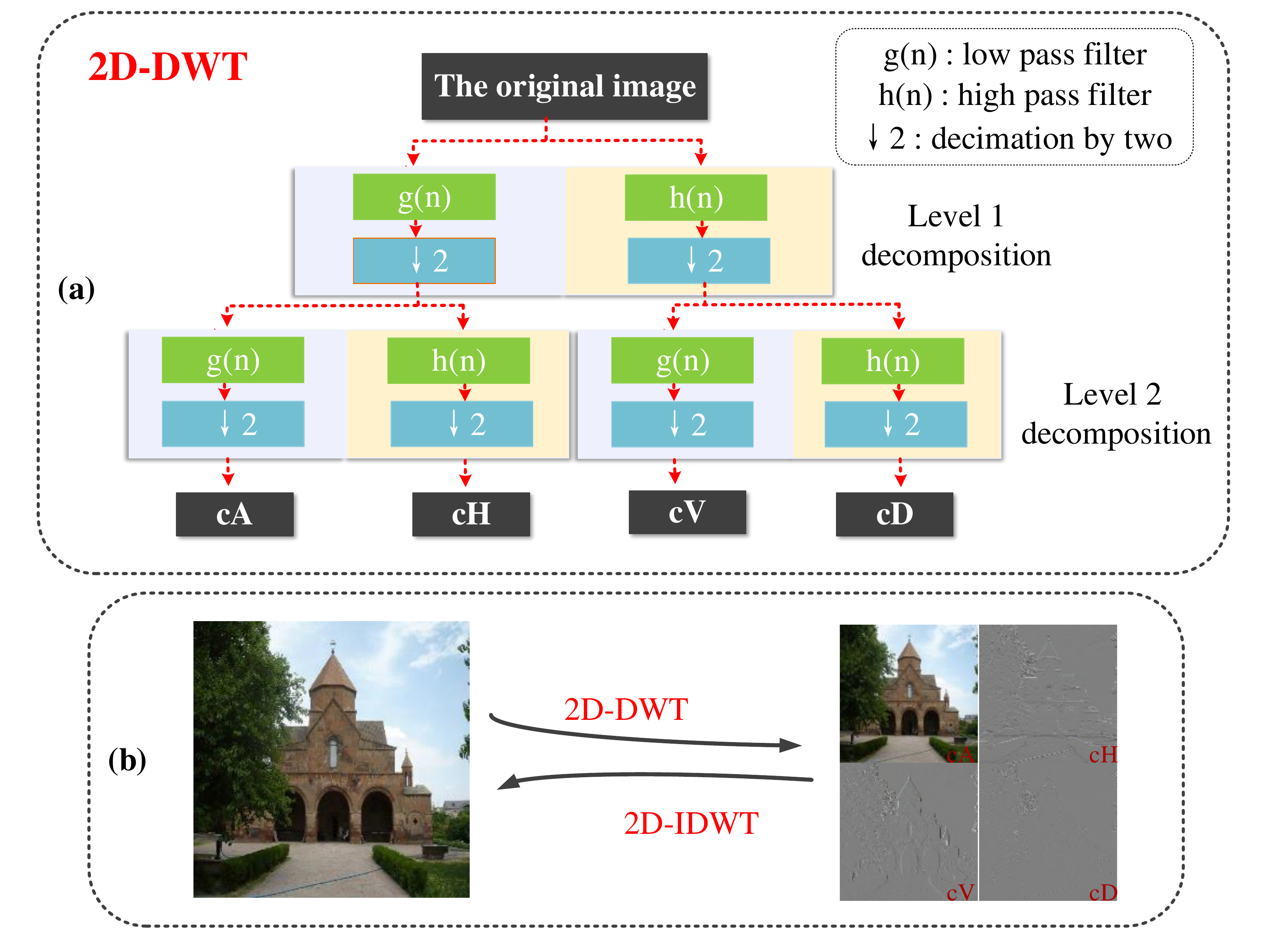}
  \vspace{-15pt}
  \caption{\hspace{-0.5em}The procedure of 2D-DWT and 2D-IDWT. (a) The flowchart of 2D-DWT. (b) Example of Haar wavelet.}\label{waveforms}
  \end{center}
  \vspace{-0.7cm}
\end{figure}

The fundamental idea behind DWT is to analyze images according to scale\cite{chowdhury2012image}, which can produce images at different frequencies. The 2D-DWT is performed by applying the 1D-DWT along the rows and columns separately and subsequently, as shown in Fig. \ref{waveforms}(a). The first analysis filter is applied to the row of the image and produces a set of approximate row coefficients and a set of detailed row coefficients. The second analysis filter is applied to the column of the new image and produces four different sub-band images, among which sub-band ${cA}$ contains approximation information of the original image. The sub-bands denoted ${cH}$, ${cV}$, and ${cD}$ contain the finest-scale detailed wavelet coefficients. Meanwhile, the 2D Inverse DWT (2D-IDWT) traces back the 2D-DWT procedure by inverting the steps, so the components can be assembled back into the original image without losing information \cite{acharya2020image}. This non-redundant image representation provides better image information compared with other multi-scale representations such as Gaussian and Laplacian pyramids. 

Typically, there are various types of wavelets, such as Haar\cite{stankovic2003haar}, Morlet\cite{lin2000feature}, Daubechies\cite{vonesch2007generalized}, etc. Different wavelets may generate various sparse representations of an image. In this study, we use the Haar wavelet to linearly decompose the image. As shown in Fig. \ref{waveforms}(b), supposing an ${2 \times 2}$ image $I=[[a,b],[c,d]]$, then the resolution of the four wavelet coefficients is ${1 \times 1}$ . The calculation process is as follows:
\begin{equation}\begin{split}\label{ideal_rectifier_resistance}
&cA=((a+b)+(c+d))/2\ \ cH=((a+b)-(c+d))/2 \\
&cV=((a-b)+(c-d))/2\ \ cD=((a-b)-(c-d))/2 .
\end{split}
\end{equation}

\subsection{DSM and Langevin Dynamics}
Generative modeling can be roughly divided into two types: explicit and implicit. GANs, as representatives of implicit generative models, adopt adversarial optimization methods and do not directly model the likelihood. Explicit generative models directly estimate the likelihood, e.g., Deep Boltzmann Machine \cite{salakhutdinov2009deep}, Variational Autoencoder \cite{kingma2013auto} and DAE, etc. Recently, Vincent \emph{et al.} \cite{vincent2011connection} proposed DSM, a variant of score matching\cite{hyvarinen2005estimation}. Defining the score of a probability density ${p(x)}$ to be the gradient of the log density ${\nabla_{x}logp(x)}$, DSM is able to estimate the score of high-dimensional data via deep networks. Furthermore,\cite{vincent2011connection} also revealed the connection that DAE is equivalent to performing DSM but DSM optimizes the data distribution more directly than DAE.

As shown in \cite{vincent2011connection,song2019generative}, defining a dataset consists of ${i.i.d.}$ samples ${\{ x_i \in \mathbb{R}^d\}^N_{i=1}}$ from an unknown data distribution ${p_{data}(x)}$, the score network ${S_\theta : \mathbb{R}^d \to \mathbb{R}^d}$ is a neural network parameterized by ${\theta}$ that will be trained to approximate the score of ${p_{data}(x)}$. DSM first perturbs the data point ${x}$ with a pre-specified noise distribution ${q_\sigma(\tilde{x}|x)}$ and then employs score matching. The objective was proved equivalent to the following:
\begin{equation}\label{related_dsm}\begin{aligned}
L_{DSM}(s)=\frac{1}{2}\mathbb{E}_{q_{\sigma}(\tilde{x}|x)p_{data}(x)}
[\lVert S_\theta(\tilde{x})-\nabla_{\tilde{x}}logq_\sigma(\tilde{x}|x)\lVert^2_2].
\end{aligned}
\end{equation}

When Eq. (\ref{related_dsm}) is minimized and the noise is small enough such that ${q_\sigma(x) \approx p_{data}(x)}$, and the optimal score network ${S_{{\theta}^*}(x)}$ satisfies ${S_{{\theta}^*}(x)=\nabla_xlogq_\sigma(x) \approx \nabla_xlogp_{data}(x)}$.

As a class of Markov Chain Monte Carlo (MCMC) techniques\cite{robert2004monte}, Langevin dynamics \cite{brooks2011handbook} provides a well-known and much studied way to sample from the distribution ${p_{data}(x)}$ using only the score function ${\nabla_xlogp_{data}(x)}$ or the trained score network ${S_{{\theta}^*}(x)}$. Langevin dynamics algorithm can be interpreted as a discrete approximation of a continuous diffusion denoted as ${Y_t}$, which started at some ${Y_0 \in \mathbb{R}^d \sim \mu_0}$. The diffusion process is given by:

\begin{equation}\label{related_dsm_1}\begin{aligned}
dY_t=\nabla logp_{\sigma}(Y_t)dt+ \sqrt{2}dB_t,
\end{aligned}
\end{equation}
where ${B_t}$ is a standard ${d}$-dimensional Brownian motion. Denoting by ${v_t}$ the law of ${Y_t}$, under quite general conditions, ${v_t}$ converges to ${p_{data}(x)}$ \cite{block2020fast,bakry2014analysis}.

\section {Proposed WACM Model}
The forward formulation of the colorization task can be mathematically expressed as:
\begin{equation}\label{basic_f}
y=F(x),
\end{equation}
where $y$ and $x$ denote the gray-level image and the original color image, $F$ denotes a degenerate function. For example, for a color image in RGB space, Eq. (\ref{basic_f}) is often considered as:
\begin{equation}\begin{split}\label{RGB}
    y=(x_R+x_G+x_B)/3.0,\\
 \end{split}
\end{equation}
or
\begin{equation}\begin{split}\label{RGB2}
    y= 0.299x_R+0.58x_G+0.114x_B.
 \end{split}
\end{equation}
The goal of colorization is to retrieve color information from a grayscale image. As discussed above, generative model has become one of the most important candidates for this task. In this study, the colorization model WACM is iterated in wavelet domain to improve the capability of score-based generative model. Specifically, WACM initially learns the prior in wavelet domain via DSM, then generates the high-quality wavelet coefficient samples via  Langevin dynamics. To further accomplish the colorization task and make synthesized color to be natural and reasonable, dual consistency terms in wavelet domain are enforced in iterations sequentially. Finally, the inverse wavelet transform is used to assemble twelve wavelet coefficients back into a high-dimensional colorized image.

\subsection{Generative Modeling in Wavelet Domain}
To advance the colorization task through the generative model with score matching, the first component in WACM is to develop a more sophisticated modeling strategy. Song \emph{et al.}\cite{song2019generative} proposed noise conditional score networks, which perturbs data with random Gaussian noise to make the data distribution more amenable to score-based generative 
modeling precisely. Let $\{\sigma_i\}^L_{i=1}$ be a positive geometric sequence that satisfies ${{{\sigma_1}/{\sigma_2}}={{\sigma_2}/{\sigma_3}}={\cdots}={{\sigma_{L-1}}/{\sigma_L}}>1}$ and ${q_\sigma(\tilde{x}|x)=N(\tilde{x}|x,\sigma^2I)}$, the unified DSM objective used in NCSN becomes:
\begin{equation}\begin{split}\label{unifiedncsn}\begin{aligned}
    L(\theta;\{\sigma\}^L_{i=1}) \triangleq   \frac{1}{2L}\sum^L_{i=1}\lambda(\sigma_i)E_{p_{data}(x)}E_{q_\sigma(\tilde{x}|x)}\\  \Vert S_\theta(\tilde{x},\sigma_i)+(\tilde{x}-x)/\sigma^2_i\Vert^2_2,
 \end{aligned}
 \end{split}
\end{equation}
where ${\lambda(\sigma_i)>0}$ is a coefficient function depending on ${\sigma_i}$. As a conical combination of DSM objectives,  ${S_\theta(x,\sigma)}$ minimizes Eq. (\ref{unifiedncsn}) if and only if ${S_\theta(x,\sigma_i)=\nabla_xlogq_{\sigma_i}(x)}$ for all ${i\in\{1,2,\cdots,L\}}$.

After ${S_\theta(x,\sigma_i)}$ is determined at the training phase, annealed Langevin dynamics as a sampling approach is introduced. It recursively computes the follows:
\begin{equation}\begin{split}\label{anealedlangevin}\begin{aligned}
    x_{t+1}&=x_{t}+\frac{\alpha_i}{2}{\nabla_x}logq_{\sigma_i}(x_t)+\sqrt{\alpha_i}z_t \\
    &=x_{t}+\frac{\alpha_i}{2}S_\theta(x_t,\sigma_i)+\sqrt{\alpha_i}z_t,\end{aligned}
 \end{split}
\end{equation}
where ${\alpha_i}$ is the step size {which gradually annealed along the geometric sequence,} $t$ is the number of iteration index for each noise level, and ${\forall t:z_t\sim N(0,I)}$. Although NCSN has achieved good results {empirically}, it still leaves a huge room for improvement, particularly in prior representation and {the dimension of manifold. } 

{The rate of convergence of Langevin dynamics is governed by a parameter of the population distribution called the log-Sobolev constant which tends to grow exponentially with the dimension of the space. And for image generation in the high-dimensional pixel space, one would expect the mixing to be so slow as to be prohibitive. However,} as the foundation of manifold learning, the manifold hypothesis states that certain high-dimensional data can be learned because they lie on or near a much lower-dimensional manifold embedded into the ambient space \cite{fefferman2016testing,belkin2003laplacian}. Block \emph{et al.}\cite{block2020fast} proved that in this paradigm, especially for highly structured data such as images, the relevant measure of the mixing of Langevin dynamics is only the intrinsic dimension of the data rather than any extrinsic features. A key conclusion is as follows: 
\begin{equation}
   {c_{LS}{(p_\sigma^{2})}=O{\left(\sigma^{2}+K^{4}{d^{'}}^{2}\kappa^{20K^{2}d^{'}}\right).}}
\end{equation}
where ${d^{'}}$ is the intrinsic dimension and ${K > 1}$, ${{\kappa} > 1}$. The above bound is completely intrinsic to the geometry of the data manifold and the dimension of the feature space does not appear. More specific proof and derivation process can be found in  \textbf{Appendix}.

\begin{figure*}[ht!]
  \begin{center}
  \includegraphics[width=7in]{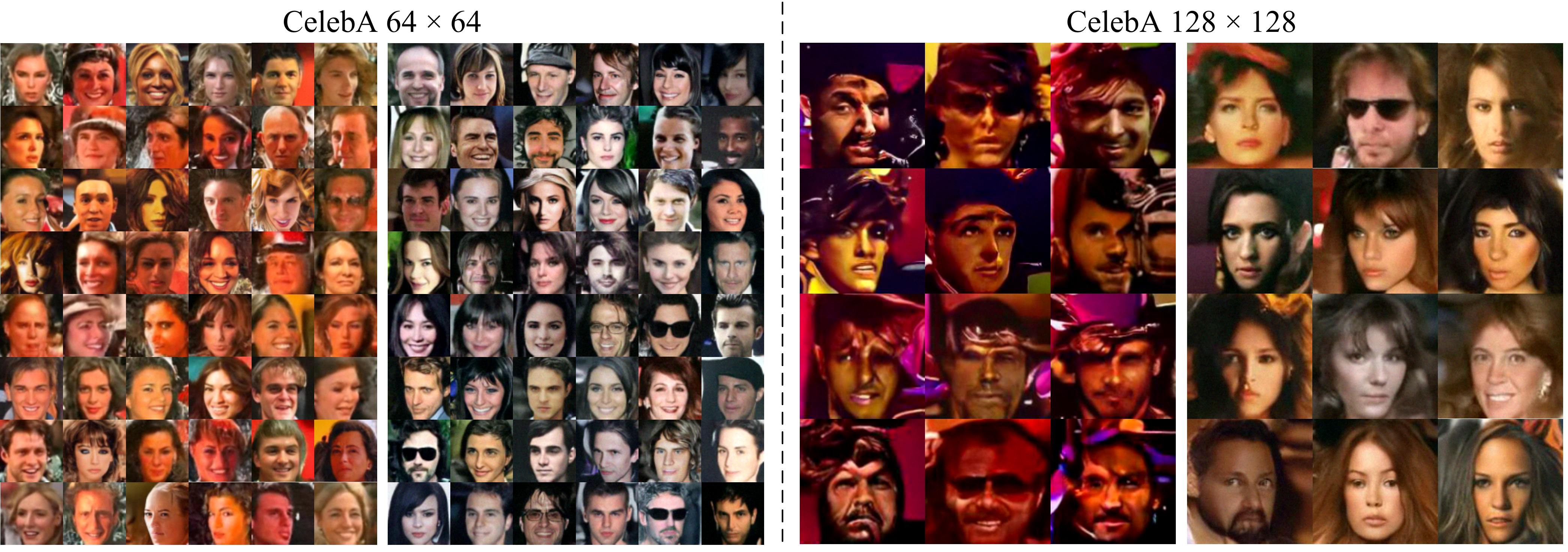}
  (a) NCSN (Iter. = 215k) \qquad\quad (b) WACM (Iter. = 160k)\qquad\quad (c) NCSN (Iter. = 295k)\qquad\quad (d) WACM (Iter. = 275k
  \vspace{-0.1cm}
  \caption{\hspace{-0.7em}{The visual results generated by the model of naïve NCSN in intensity domain and WACM in wavelet domain. Left: ${64 \times 64}$, right: ${128 \times 128}$. It can be observed that naïve NCSN performs fair against our model in ${64 \times 64}$ images. Both of them generate appropriate and realistic results in (a)(b). However, influenced by the curse of dimension, naïve NCSN is not capable {of generating} complete and clear results in higher resolution images with size of ${128 \times 128}$ and only generates chaos images with basic features of human faces as illustrated in (c). By contrast, {benefiting} from the proposed method, our results perform excellently in ${128 \times 128}$ images, which is significantly better than the naïve NCSN. This phenomenon strongly indicates the superiority and effectiveness of this strategy.
}}\label{naive compare}
  \end{center}
  \vspace{-0.6cm}
\end{figure*}

\begin{figure}[ht!]
  \begin{center}
  \includegraphics[width=3.5in]{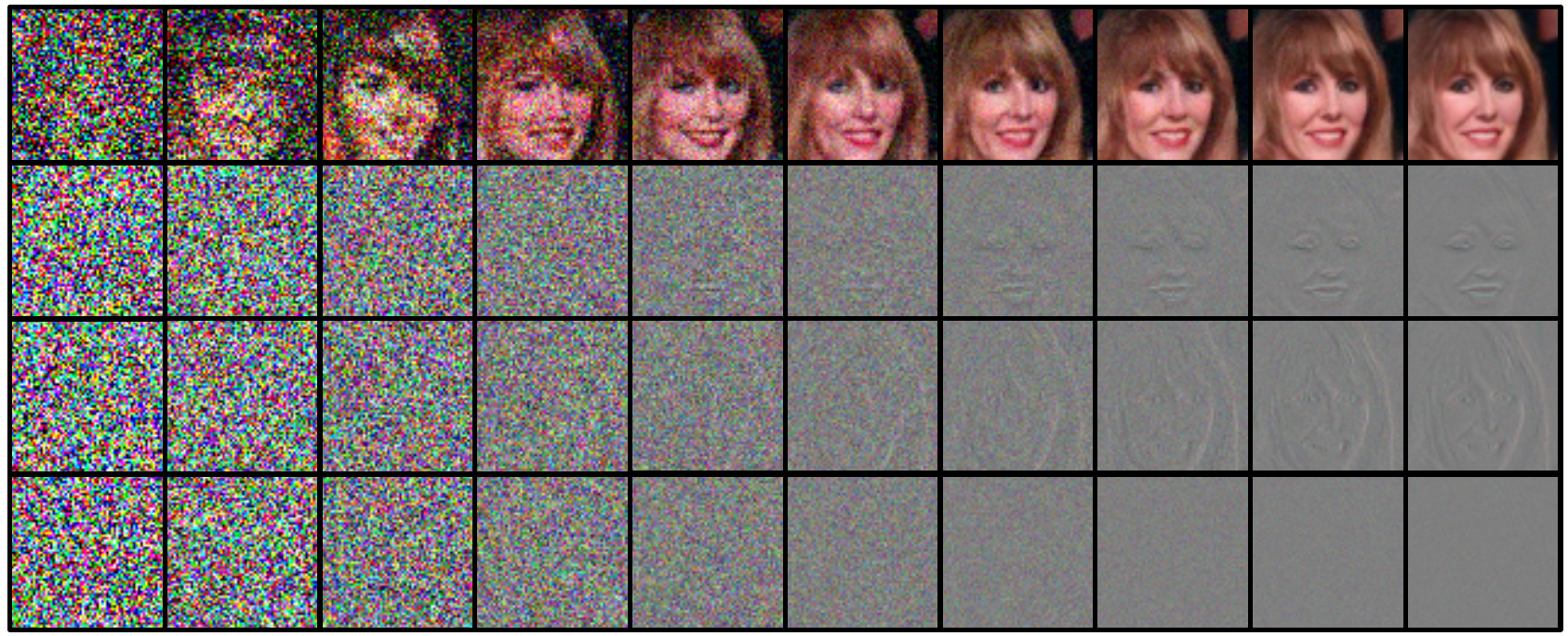}
  \vspace{-0.4cm}
  \caption{\hspace{-0.4em}{Sampling trajectories of the wavelet coefficients. Notice that the low-frequency component mixes at an earlier stage (i.e., the fifth column). A}t the same time, the other high-frequency components mix more slowly (i.e., the seventh column).}\label{iteration}
  \end{center}
  \vspace{-0.7cm}
\end{figure}

Despite the intrinsic dimension is much smaller than the apparent dimension, the authors of \cite{block2020fast} still argued that the high dimension of the sampling space significantly impairs the performance of Langevin dynamics, and there can be very bad dimensional dependence in the score estimation. Furthermore, training and sampling in high dimension are also compute intensive. Following the above {analysis}, the authors in \cite{block2020fast} proposed a multi-resolution strategy based on upsampling to reduce the dimension of data, which can transfer some of the hard work of score estimation to an easier, lower-dimensional regime. This way falls into the progressive strategy, {that is,} the images are generated from low-dimensional resolution to high-dimensional resolution progressively. But unfortunately, compared with the naïve NCSN, experiments demonstrated that the progressive strategy utilizing the multi-resolution scheme lacked significant improvement.

Our idea is also motivated by {the theoretical analysis to data dimension above, while achieves a better performance in practice.} Different from the “sequence” method in \cite{block2020fast}, we take advantage of the wavelet transform in a “joint” manner by {processing image into multi-scale and multi-channel representation, thus to make the score-based generative model estimating and sampling in a lower-dimensional but more informative wavelet subspace}. 

Intuitively, the multi-scale decomposing and downsampling processing operations on images in wavelet transform allow some hard parts of generative modeling to be easily transferred to a lower-dimensional subspace without adding strong conditions on the target density \cite{sutherland2018efficient}. Furthermore, compared to the original image, this representation contains one low-frequency sub-band image with complex content and three high-frequency sub-band images with simple content, which provides richer statistical information for the model to learn more favorable priors and improve capability. In addition, thanks to the IDWT process, the generated wavelet coefficients can be assembled back into the image of original resolution non-destructively and almost without spending extra time, which also reduces computational complexity and speeds up the runtime compared with the "sequence" method.

More concretely, supposing ${x}$ is a target image containing the three color-channel of ${R,G,B}$, which can be expressed as ${x=[x_R,x_G,x_B]}$. Applying DWT to each channel, it yields
\begin{equation}\begin{split}\label{DWT_RGB}\begin{aligned}
    W(x_R)=[cA_R,cH_R,cV_R,cD_R]=W_R \\
    W(x_G)=[cA_G,cH_G,cV_G,cD_G]=W_G \\
    W(x_B)=[cA_B,cH_B,cV_B,cD_B]=W_B,
    \end{aligned}
 \end{split}
\end{equation}
where ${W_R, W_G, W_B}$ are three four-channel tensors superimposed by the four sub-band images whose resolution is one-quarter of the reference.

Stacking the three tensors together, a 12-channel tensor ${X=[W_R, W_G, W_B]}$ is obtained to train the network. The goal of stacking to be ${X}$ is to form object in multiple lower-dimensional manifold jointly in favor of the subsequent network learning \cite{quan2021homotopic,liu2020highly}, thereby avoiding potential difficulties with both accuracy in score estimation and sampling with Langevin dynamics. Accordingly, the objective of WACM is:
\begin{equation}\begin{split}\label{X_unified ncsn}\begin{aligned}
    L(\theta;\{\sigma\}^L_{i=1}) \triangleq \frac{1}{2L}\sum^L_{i=1}\lambda(\sigma_i)E_{p_{data}(x)}E_{q_\sigma(\tilde{X}|X)}\\  \Vert S_\theta(\tilde{X},\sigma_i)+(\tilde{X}-X)/\sigma^2_i\Vert^2_2.
 \end{aligned}
 \end{split}
\end{equation}

To investigate the multi-scale and joint-learning strategy of WACM, we train the naïve NCSN and WACM on CelebA dataset in ${64 \times 64}$ and ${128 \times 128}$, respectively. The intermediate generated results of modeling in wavelet domain are shown in Fig. \ref{iteration}. It can be observed that, as the iteration increases, the intermediate results approach the ground truth gradually. The 
low-frequency component mixes at an earlier stage (i.e., the fifth column), meanwhile, the other high-frequency components mix more slowly (i.e., the seventh column).

The generation comparison between the results of modeling in intensity or wavelet domain is shown in Fig. \ref{naive compare}. The generation effect of WACM is significantly better than the naïve NCSN for CelebA ${128 \times 128}$. In addition, because the face position of the CelebA data set is aligned and the face images are smooth. The data distribution of the high-frequency wavelet coefficients is relatively regular and the network can learn the prior and generate the subspace information faithfully.

\begin{figure*}[ht!]
  \begin{center}
  \includegraphics[width=7in]{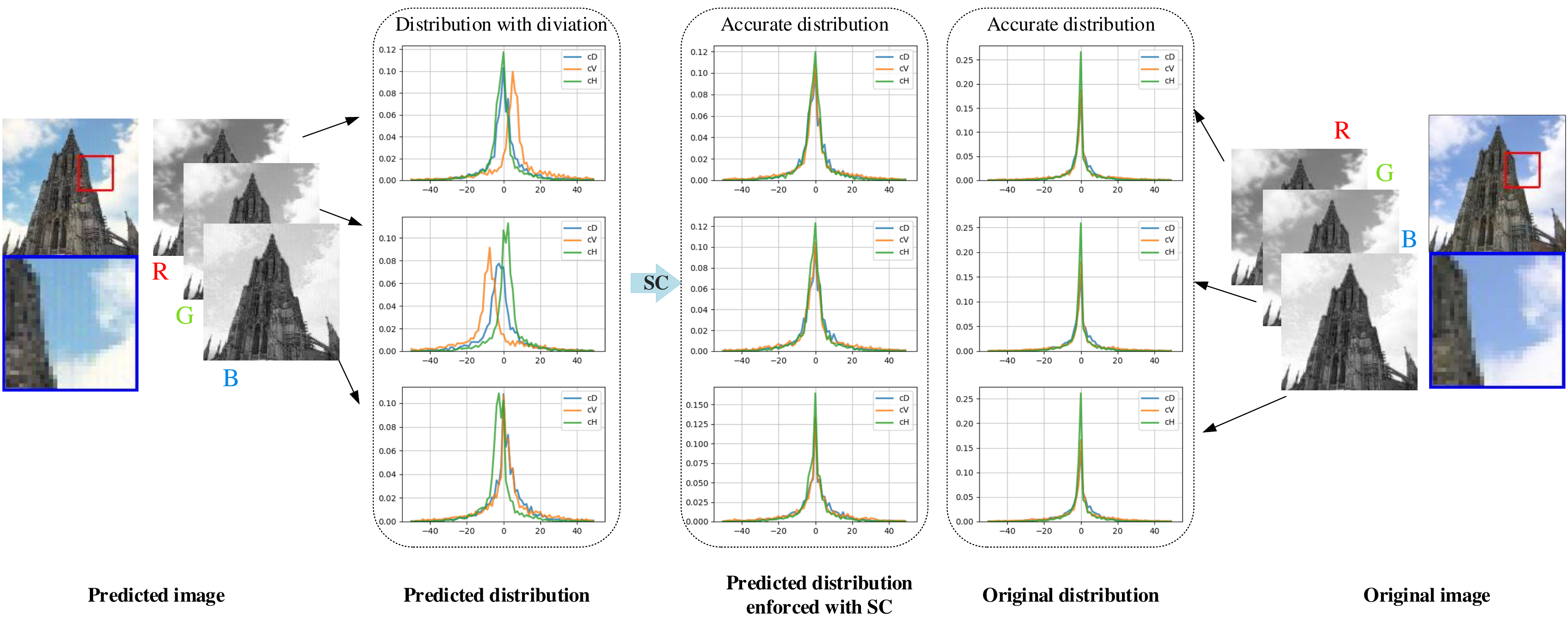}
   \vspace{-0.1cm}
  \caption{\hspace{-0.5em}{The histograms of the high-frequency wavelet coefficients of the R, G, and B channels of the original image (a) and the predicted image (b). Compared 
with the original image, the high-frequency histograms of the generative image have dissimilar distributions, which leads to errors in the edge and "gridding" effects of the generated image.
}}\label{mc_motivation}
  \end{center}
  \vspace{-0.6cm}
\end{figure*}

\begin{figure}[ht!]
  \begin{center}
  \includegraphics[width=3.5in]{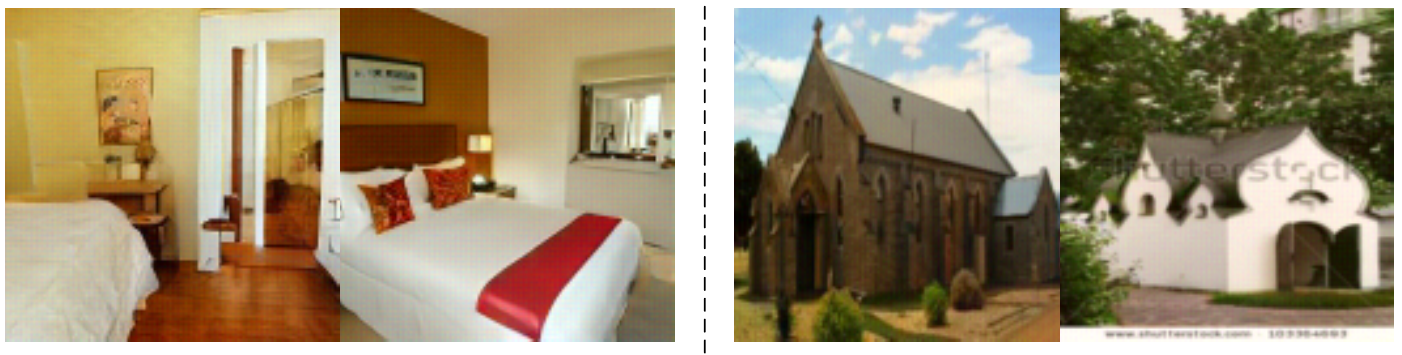}
   \vspace{-0.15cm}
  \small{(a)\qquad \hspace{+4cm}  (b)}
  \caption{\hspace{-0.6em}{Results generated by WACM with DC term. (a) Colorized bedrooms with DC. (b) Colorized churches with DC. Although the colorization results are natural overall, they suffer from improper gridding effects in detail.}}\label{resultswithdc}
  \end{center}
  \vspace{-0.75cm}
\end{figure}
\subsection{{Colorization under Two Consistencies}}
The key to {utilizing} score-based generative model for colorization and {reducing} the intrinsic limitations lies in the design of proper consistency strategies. Consequently, in the second 
component of WACM, data-consistency and structure-consistency are devised to guide the model to achieve superior colorization performance.

\subsubsection{\textbf{Data-Consistency in Wavelet Domain}} To limit the unconditional generative model and guide it to colorize the input grayscale image, a Data-Consistency (DC) term is proposed and added in the iterative procedure. More precisely, the DC term guides the generative model to complete the colorization task on the input grayscale by minimizing the error between the observed value of the intermediate result at each iteration and the sub-band image of the original input.  

Because of the linear relationship between the degenerate 
function ${F}$ and the Haar wavelet ${W}$, the order of the two operations is commutative. Thus, the following equation can be obtained as:
\begin{equation}\label{change_f}\begin{aligned}
   W(y)=W(F(x))=F(W(x)),
 \end{aligned}
\end{equation}
and
\begin{equation}\begin{split}\label{x}\begin{aligned}
    & cA_y=F(cA_R,cA_G,cA_B) \ cH_y=F(cH_R,cH_G,cH_B) \\
    & cV_y=F(cV_R,cV_G,cV_B)\ \ \ cD_y=F(cD_R,cD_G,cD_B).
 \end{aligned}
 \end{split}
\end{equation}

Therefore, the DC term can be directly applied to the wavelet domain, that is, the 12 channels of ${X}$ as:
\begin{equation}\begin{split}\label{DC}\begin{aligned}
    {w_1DC(X)}&{=w_1(F(W(x))-F(y))} \\
    &{=w_1(F(X)-F(y))},
 \end{aligned}
 \end{split}
\end{equation}
where {${w_1}$} is a hyper-parameter that is related to the noise level at the current iteration.

The colorization results of WACM with only data-consistency in wavelet domain are shown in Fig. \ref{resultswithdc}. It {demonstrates} that, after the DC term is enforced, the model {can already} perform basic colorization on the input grayscale image, but due to the deviation of the generated wavelet coefficients, the final effect still has certain structural defects.

\subsubsection{\textbf{Structure-Consistency in Wavelet Domain}}The proposal of the Structure-Consistency (SC) is based on the observation of the overall RGB color deviation and grid phenomenon in the colorization results after the DC term is applied.

As shown in Fig. \ref{mc_motivation}, we output the histograms of the high-frequency wavelet coefficients of the R, G, and B channels of the original color picture and the gridded picture, respectively. Due to the inherent freedom of the generative model, the data distribution of the obtained ${cH}$, ${cV}$, and ${cD}$ have certain deviations compared with the original RGB image. Since the IDWT result is very sensitive to the wavelet coefficients, especially the high frequency components, these deviations will cause display defects of edge differences and grid phenomenon in the final colorization results.

{It can be observed that the high-frequency wavelet distribution curve of the RGB channel and of the grayscale are similar to the normal distribution without outliers, and the mean or median of the distribution can be used to approximate the center. On this basis,} we devise the SC term to correct the generated high-frequency wavelet coefficients using the mean of the grayscale input. For the ${i}$-th channel ${X_i}$ in ${X}$, SC can be expressed as:
\begin{equation}\begin{split}\label{SC}\begin{aligned}
    {w_2SC(X_i)=w_2(Mean(X_i)-Mean(W(y)_i))}.
 \end{aligned}
 \end{split}
\end{equation}
{where ${w_2}$ is the weight of SC which is set to 1 by default.} For each channel of {high-frequency wavelet coefficients}, the ${SC(X_i)}$ is the difference between the mean value of the channel and the mean value of the corresponding wavelet coefficient of the input grayscale image. The calculated ${SC(X_i)}$ of each channel is a real number, and then ${X_i}$ is modified by subtracting a product of the difference and ${w_2}$ from each pixel.

After each iteration, SC is used to correct the {high-frequency wavelet coefficients of intermediate result by shifting the coefficients as a whole to the correct distribution. Notably, the SC term is only applied to high frequency wavelet coefficients. Otherwise, applying SC to the low-frequency wavelet coefficients will result in the colorization with low saturation, because this will make the distribution center of RGB channels tend to the same position, and the same RGB value visually appears as gray.}

\begin{figure*}[t]
  \begin{center}
  \includegraphics[width=7in]{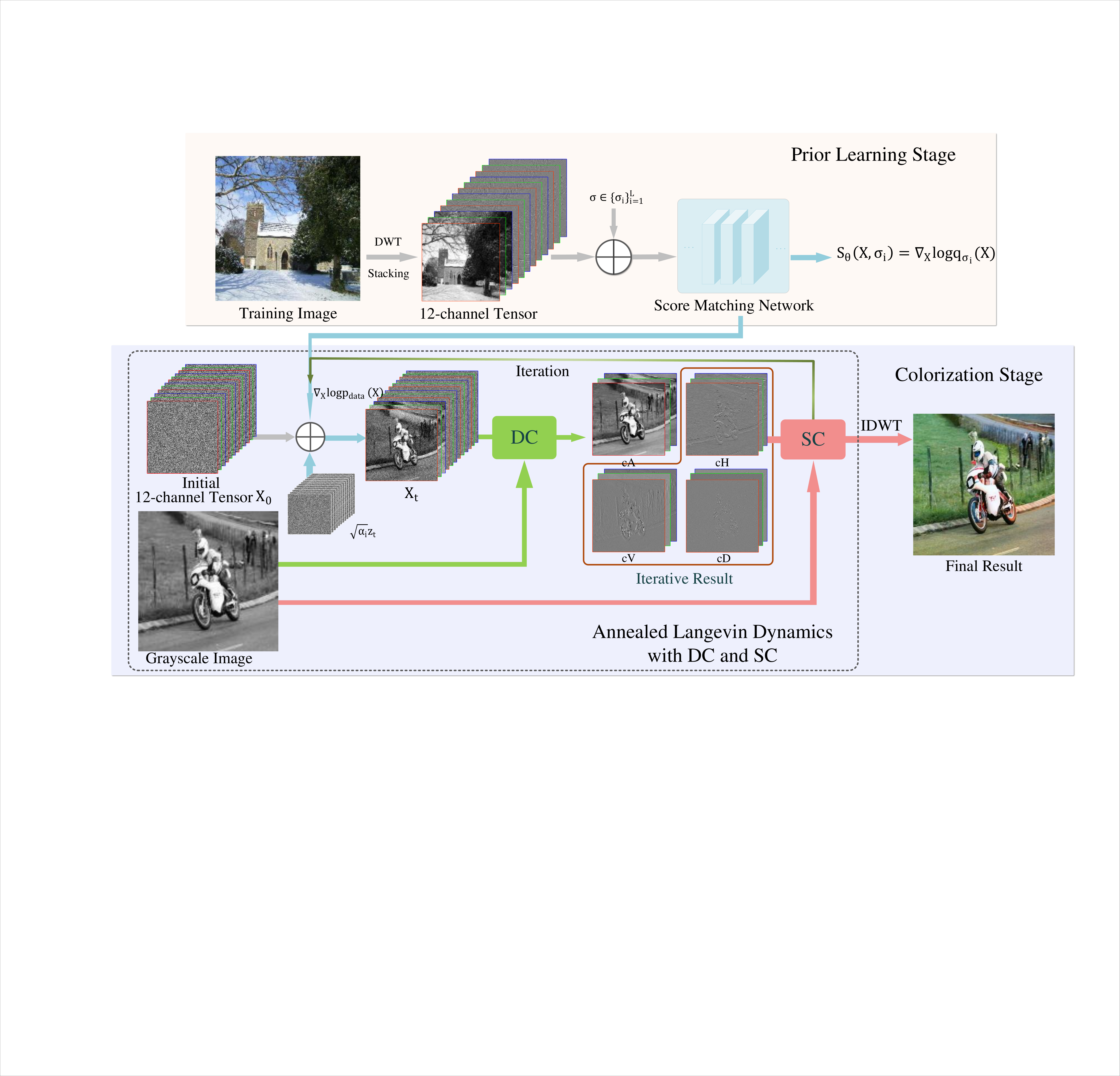}
  \vspace{-0.2cm}
  \caption{\hspace{-0.6em}{Iterative colorization procedure of WACM. Specifically, in the prior learning stage, the network ${S_\theta(X)}$ learns to retrieve ${\nabla_Xlogp_{data}(X)}$ in wavelet 
domain that best matches the ground truth of the input image. In the colorization stage, WACM generates samples from the 12-dimensional noisy data distribution by annealed Langevin dynamics with data-consistency. At the meantime, the structure-consistency is used to improve the performance and reduce the improper effects of the samples. Here symbol ${"\bigoplus"}$ stands for the sum operator, DC and SC stand for data-consistency and structure-consistency.
}}\label{structure}
  \end{center}
  \vspace{-0.7cm}
\end{figure*}

\subsection{Summary of WACM}

With the above-mentioned dual consistency terms, the model can better utilize the wavelet transform in the colorization task with score matching. Overall, as Fig. \ref{structure}, the entire colorization diagram includes two processes: learning prior information in wavelet domain and iterative generate colorization process.
Specifically, in the training phase, a set of 12-channel tensors {is} formed by applying wavelet transform to the R, G, B channels of an image respectively to train the DSM network in the multiple low-dimensional space. After the network is trained, the model can sample with the annealed Langevin dynamics which recursively computes the following formula modified with the DC term:
\begin{equation}\begin{split}\label{DC}\begin{aligned}
    X_{t+1}=X_t+\frac{\alpha_i}{2}S_\theta(X_t,\sigma_i)-w_1 DC(X_t)+\sqrt{\alpha_i}z_t,
 \end{aligned}
 \end{split}
\end{equation}
where ${\forall t:z_t \sim N(0,I)}$.

{In the sampling process, a 12-channel tensor ${X_0}$ is initialized from uniform noise as the input for the first iteration, and a list of noise levels ${\{\sigma_i\}^L_{i=1}}$ which is reduced proportionally, is generated for each step of the outer loop. At each iteration, the annealed Langevin dynamics samples an intermediate result from ${q_{\sigma}(X)}$, and then artificial noise is added to the intermediate result based on the noise level ${\sigma_i}$. This transition helps smoothly transfer the benefits of large noise levels to low noise levels where the perturbed data is almost indistinguishable from the original one. At the same time, the DC is incorporated into the update of the iterative generative process, which constitutes the inner loop of the iterative generation step jointly. Then, the proposed SC is applied to the generated wavelet coefficients in the outer loop. When ${\sigma_L\approx0}$, ${q_{\sigma}(X)}$ is close to ${p_{data}(X)}$, the results of color channels expressed in wavelet coefficients are obtained. Finally, the final result can be attained by performing an inverse wavelet transform on the iteratively generated wavelet coefficients. The whole sampling process is explained in \textbf {Algorithm 1}.}
\begin{table}[H]
\normalsize
\vspace{-0.1cm}
\renewcommand{\arraystretch}{1.0}
\begin{center}
\begin{tabular}{l}
\hline\hline
\makecell[c]{\textbf{\leftline{Algorithm 1. Iterative Colorization via WACM}}} \\
\hline\hline

\textbf{Initialization:} \\
(a) Set parameters ${\varepsilon}$, ${L}$, ${T}$, ${w_1}$ and ${w_2}$. \\
(b) Initialize noise level ${\{\sigma_i\}^L_{i=1}}$ and ${X_0 \sim U(-1,1)}$.\\

\textbf{Outer loop:}\\ 
\textbf{For} ${i=1,2,\cdots,L}$ \textbf{do}\\
\hspace{0.4cm}(c) Set the step size ${\alpha_i}=\varepsilon \cdot \sigma^2_i / \sigma^2_L$.\\
\hspace{0.4cm}\textbf{Inner loop:} \\
\hspace{0.4cm}\textbf{For} ${t=1,2,\cdots,T}$ \textbf{do} \\
\hspace{0.8cm}(d) ${X_{t+1}=X_t+\frac{\alpha_i}{2}S_\theta(X_t,\sigma_i)-w_1 DC(X_t)+\sqrt{\alpha_i}z_t}$\\

\hspace{0.4cm}\textbf{End For} \\
\hspace{0.4cm}(e) Calculate ${SC(X_T)}$ via Eq. (\ref{SC}). \\
\hspace{0.4cm}(f) Update ${X_T=X_T-w_2SC(X_T)}$. \\
\hspace{0.4cm}(g) Output the colorization result  ${x=IDWT(X_T)}$. \\
\textbf{End For} \\
\hline\hline
\end{tabular}
\end{center}
\vspace{0.1cm}
\end{table}

\section{Experiments}
In this section, after the experimental setup is detailed, the present WACM is compared with the state-of-the-arts qualitatively and quantitatively. Then, several key factors that contribute to the final WACM are separately investigated. In addition, two main advantages of WACM are exhibited: colorization robustness and diversity. Finally, the effects of pre-processing and post-processing on the colorization results and some existing limitations in real-world applications are discussed as well. For the purpose of replicating research, the code is available at: \href{https://github.com/yqx7150/WACM}.

\begin{table*}[]
\small
\renewcommand{\arraystretch}{1.2}
    \centering
     \caption{ Colorization comparison of our model to state-of-the-art techniques in the ${128 \times 128}$ images.}
     \vspace{-0.2cm}
    \begin{tabular}{c|c|c|c|c}
    \hline\hline
    \makecell[c]{\textbf{Algorithm}} & \makecell[c]{\textbf{LSUN-church}} & \makecell[c]{\textbf{LSUN-bedroom}} & \textbf{COCO-stuff} & \textbf{Runtime(s)} \\ 
\hline\hline
Zhang \emph{et al.}(2016) & 23.65/0.9228 & 20.89/0.8946 & 20.21/0.8844 & 1.896\\
\hline
MemoPainter & 21.66/0.8767 & 22.92/0.8975  & 22.05/\textbf{0.8929} & 0.244\\
\hline
ChromaGAN & 24.63/0.9106 & 24.16/0.8899 & \textbf{22.98}/0.8924 & 0.378\\
\hline
iGM-6C & 20.60/0.8953 & 22.40/\textbf{0.9099} & 19.68/0.8493 & 0.055/iter\\
\hline
WACM & \textbf{25.44}/\textbf{0.9265} & \textbf{24.13}/0.9056 & 22.41/0.8810 & \textbf{0.044/iter} \\
\hline\hline    
    \end{tabular}
    \label{cp_128_table}
    \vspace{-0.5cm}
\end{table*}

\subsection{Experiment Setup}
\subsubsection{\textbf{Datasets}}We experiment with multiple image datasets from various sources as follows:

LSUN \cite{yu2015lsun} (bedroom and church): LSUN contains around one million labeled images for each of 10 scene categories and 20 object categories, including bedroom, fixed room, living room, classroom, church, etc. In this study, we choose the indoor scene LSUN-bedroom 
dataset and the outdoor scene LSUN-church dataset to evaluate WACM.

COCO-stuff \cite{caesar2018coco}: The COCO-stuff is a subset of the COCO dataset \cite{lin2014microsoft} generated for scene parsing. It contains 164k images that span over 172 categories, including 80 
things, 91 stuff, and 1 class unlabeled, most of which are natural scenes with various objects. 

\subsubsection{\textbf{Implementation Details}}{Following \cite{song2019generative}, the proposed WACM selects the 4-cascaded RefineNet \cite{lin2017refinenet} architectures with instance normalization and dilated convolutions as the score network, and continue to use the hyperparameters in \cite{lin2017refinenet,song2019generative}.} Adam is chosen as the optimizer with a learning rate of 0.005 and ${\beta1}$ of 0.9. For the setting of parameters, we choose ${L=10}$, ${\sigma_1=1}$, ${\sigma_{L}=0.01}$ to determine the noise level, ${T=100}$ as the total number of iterations for each level. Besides, ${w_1}$, ${w_2}$ and step size are chosen to be ${w_1=\alpha_i/{\sigma_i}^2}$, ${w_2=1}$ and ${\varepsilon=1.56\times10^{-5}}$ via the grid search method of hyperparameter optimization. As shown in Fig. \ref{app} in \textbf{Appendix}, the colorization results are sensitive to the parameters. In the training phase, we reshape each image into ${128 \times 128}$ or ${256 \times 256}$, then perform random flip as pre-processing. In each dataset, the WACM model is trained for 500,000 iterations with a batch size of 8, and checkpoints are saved every 5000 iterations, which takes around 40 hours. The model is performed with Pytorch interface on 2 NVIDIA Titan XP GPUs, 12 GB RAM. In the testing phase, we randomly choose 100 images from the validation set for each dataset, then 12 diverse results are produced for each grayscale image.

\subsubsection{\textbf{Evaluation Metrics}} Two quantitative assessments of our method are included in terms of Peak Signal to Noise Ratio (PSNR) and Structural Similarity Index Measure (SSIM). In addition, Lightness-Order-Error (LOE)\cite{wang2013naturalness} and a user study are conducted to test the naturalness of different methods.

\subsection{Comparisons with State-of-the-arts}
To demonstrate the superiority of the proposed WACM, we compare it with four state-of-the-art colorization methods quantitatively and qualitatively, including Zhang \emph{et al.}(2016)\cite{zhang2016colorful}, MemoPainter\cite{yoo2019coloring}, ChromaGAN \cite{vitoria2020chromagan} and iGM-6C \cite{zhou2020progressive}. Meanwhile, to further test the colorization quality of WACM on higher resolution images, we qualitatively compare WACM with mGANprior \cite{gu2020image}, which reports excellent colorization results on LUSN-bedroom and LSUN-church datasets by using Multi-Code GAN Prior.

\subsubsection{\textbf{Quantitative Comparison}} 
\begin{figure*}[ht!]
  \begin{center}
  \includegraphics[width=6.8in]{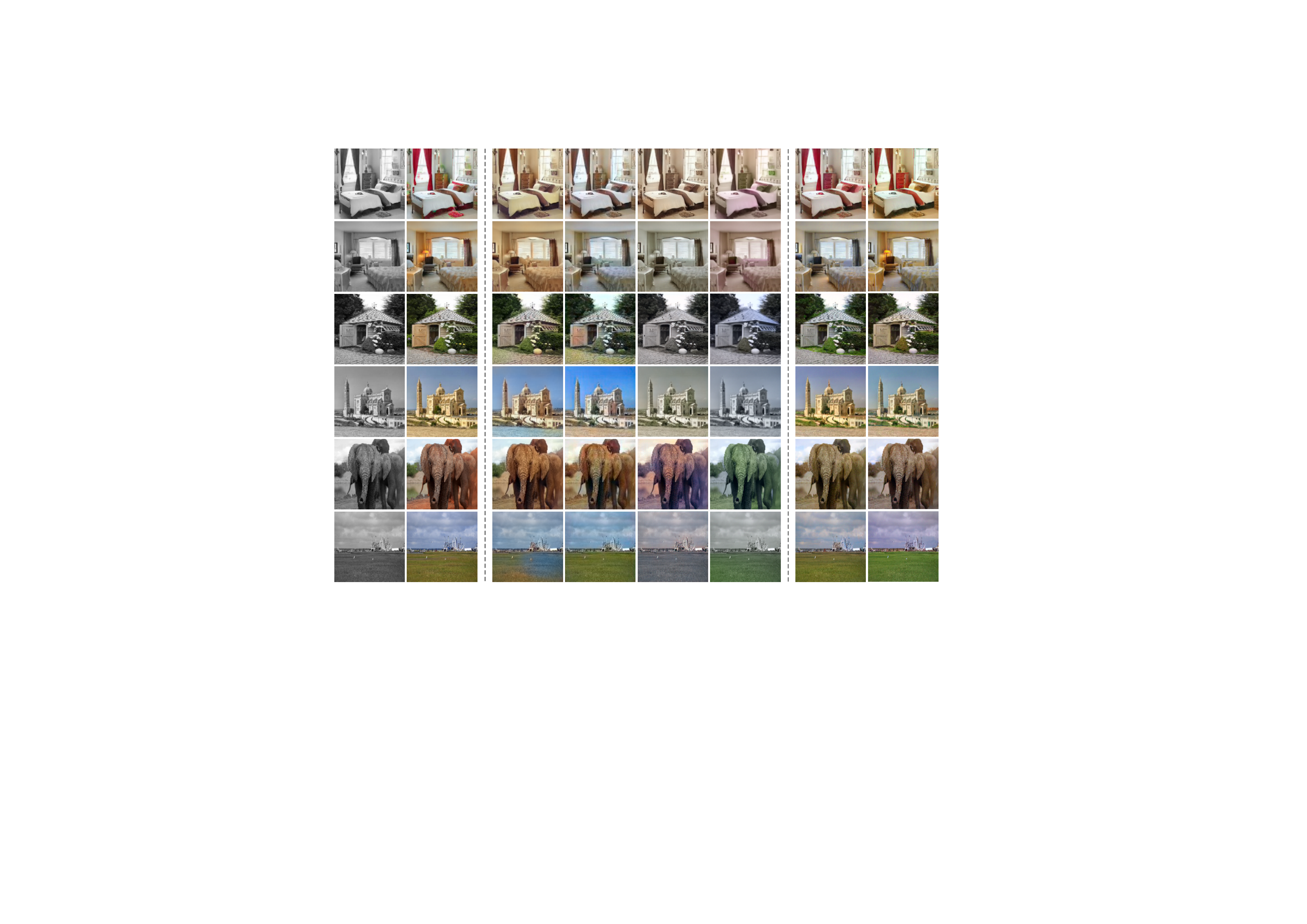}
  \leftline{\footnotesize{\quad \ \ \ (a) Grayscale \ \hspace{0.1cm} (b) Ground truth \qquad (c) Zhang \emph{et al.} \ (d) ChromaGAN  (e) MemoPainter \ \hspace{0.2cm} (f) iGM-6C \qquad \ \ \ (g) WACM \qquad\hspace{0.1cm}   (h) WACM}}
  \vspace{-15pt}
  \caption{\hspace{-1em}{Visual comparisons with the state-of-the-arts on images with the size of ${128 \times 128}$. From left to right: Grayscale, Ground truth, Zhang \emph{et al.}, ChromaGAN, MemoPainter, iGM and two diversity results of WACM. The present WACM can predict more visually pleasing colors.
}}\label{pics_cp_128}
  \end{center}
  \vspace{-0.3cm}
\end{figure*}

In this experiment, we randomly select 100 images from LSUN-bedroom, LSUN-church, and COCO-stuff datasets, respectively, and resize them to be 
${128 \times 128}$, then calculate the average PSNR and SSIM values of the results that are colorized by different methods. Table \ref{cp_128_table} and Fig. \ref{pics_cp_128} summarize the colorization performance of WACM and other state-of-the-art methods on ${128 \times 128}$ images.

One can observe that, in general, the PSNR and SSIM values of WACM are higher than most of those obtained by other methods. In LSUN-church dataset, WACM achieves the highest PSNR and SSIM values, as well as the highest PSNR values in LSUN-bedroom dataset. For COCO-sutff dataset which consists of more complex outdoor images, the ability of generative model is limited to a certain extent.However, WACM still represents strong colorization performance with the help of the multi-scale and multi-channel strategies, and the value of PSNR is slightly lower than that of ChromaGAN.

\begin{figure*}[ht!]
  \begin{center}
  \includegraphics[width=6.9in]{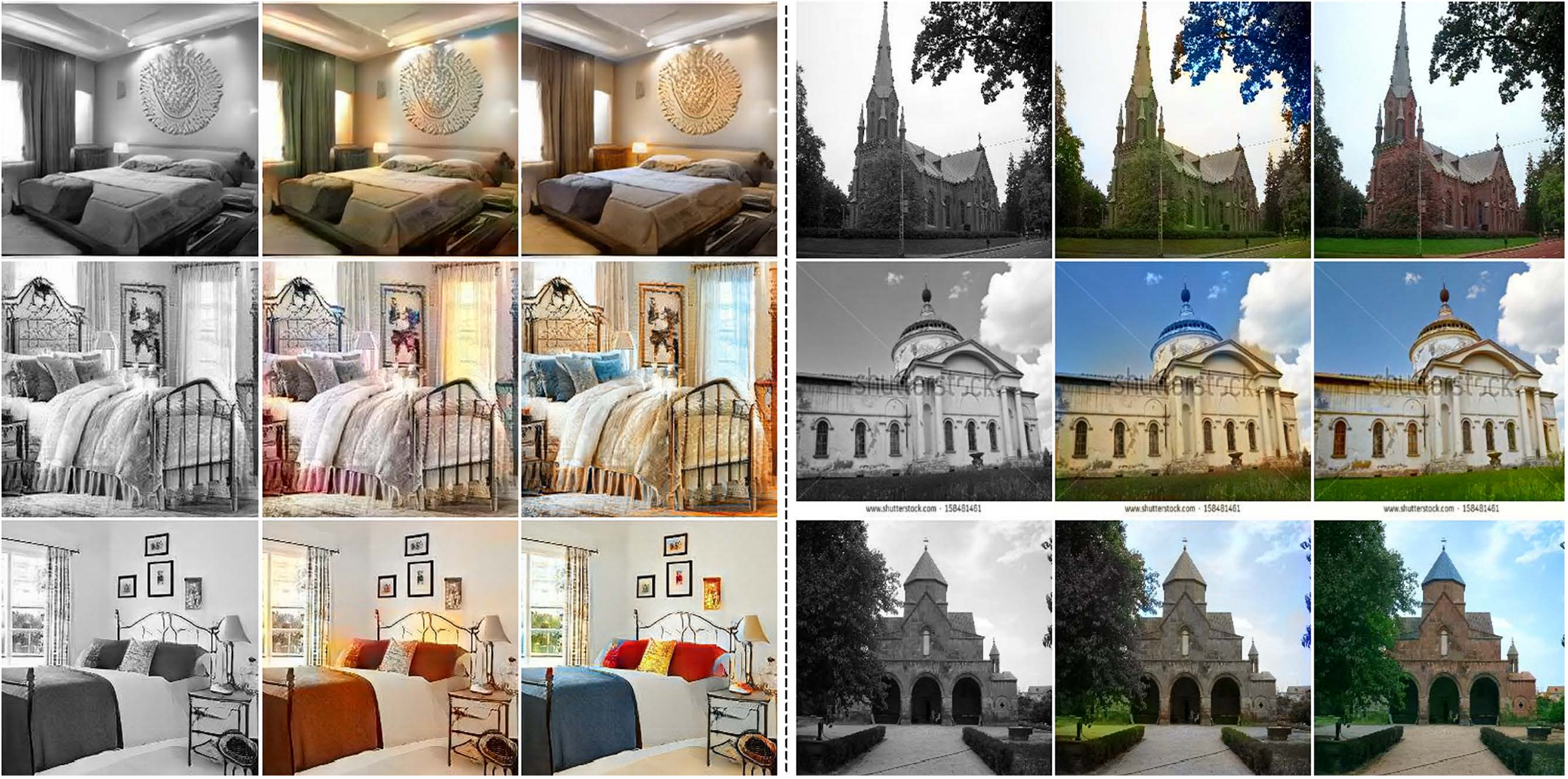}
   \vspace{-0.1cm}
  \leftline{\small{\hspace{0.8cm} (a) Grayscale \hspace{0.8cm} (b) mGANprior \hspace{0.8cm}  (c) WACM \hspace{0.8cm} \hspace{0.5cm} (a) Grayscale \hspace{0.8cm} (b) mGANprior \hspace{0.7cm} \hspace{0.5cm} (c) WACM  }}
  \caption{\hspace{-0.5em}{Some colorization results of mGANprior and WACM in ${256 \times 256}$ images. The images in the first column are the input grayscales, and the images in the second and third column are results of mGANprior and WACM, respectively. Benefiting from the multi-scale and multi-channel characteristics of DWT, WACM produces high-quality colorization 
results in higher resolution images, which alleviates the difficulty of NCSN to generate high-resolution images. 
}}\label{pics_cp_256}
  \end{center}
  \vspace{-0.5cm}
\end{figure*}

For the sake of comparison, some results are depicted in Fig. \ref{pics_cp_128}. Overall, the results of other methods provide sometimes vivid colors as in the second line and sixth line in Fig. \ref{pics_cp_128}(d) and sometimes uncolored results as in the sixth line in Fig. \ref{pics_cp_128}(c)(e). However, their results suffer from the issues of color pollution and desaturation. On the contrary, WACM yields better results in terms of consistent hue, saturation, and contrast, etc. For example, in the third row of Fig. \ref{pics_cp_128}(g), there are no discordant green colors on the ground like (c) and (d), and the image of WACM in the second row has obvious contrast in luminance between table lamp and the bed. The quantitative comparison to state-of-the-art methods indicates the superiority of WACM in aspects of naturalness and structural characteristics, including luminance, contrast, and structure.

Furthermore, to prove the contribution of wavelet transforms to the colorization performance of higher resolution images, the colorization results of WACM and mGANprior \cite{gu2020image} on ${256 \times 256}$ images are shown in Fig. \ref{pics_cp_256}. It can be observed that the results of WACM are quite realistic and free of color pollution. The results further illustrate the superiority of WACM in promoting colorization task on higher resolution images by integrating multi-scale and multi-channel strategies with score-based generative model. {More diverse examples of WACM are included in \textbf{Supplementary Materials}.}

\subsubsection{{\textbf{Naturalness Test}}}{We use the LOE indicator and a user survey to test the naturalness of colorization results accurately.} 

{LOE is an indicator by measuring the relative order of illumination between the original image and the enhanced image to measure its naturalness. The smaller the error, the better the naturalness. In this study, we use it as an indicator of the preservation of naturalness between the grayscale image and the colorized image. The results are shown in Table \ref{LOE}, which proves that the colorization results of WACM are real and natural.}

Similar to \cite{fang2019superpixel}, the user study is designed using the Two-Alternative Forced Choice (2AFC) paradigm. We choose five random colorization results generated by four methods (ChromaGAN, MemoPainter, Zhang \emph{et al.} and WACM) to make the comparison and invite 68 users in different age groups to participate in this study. For each  image, there are 6 pairs of colorized results to make sure any two methods are compared. The order of image pairs is randomized to avoid bias. 

\begin{figure}[]
  \begin{center}
  \includegraphics[width=3.4in]{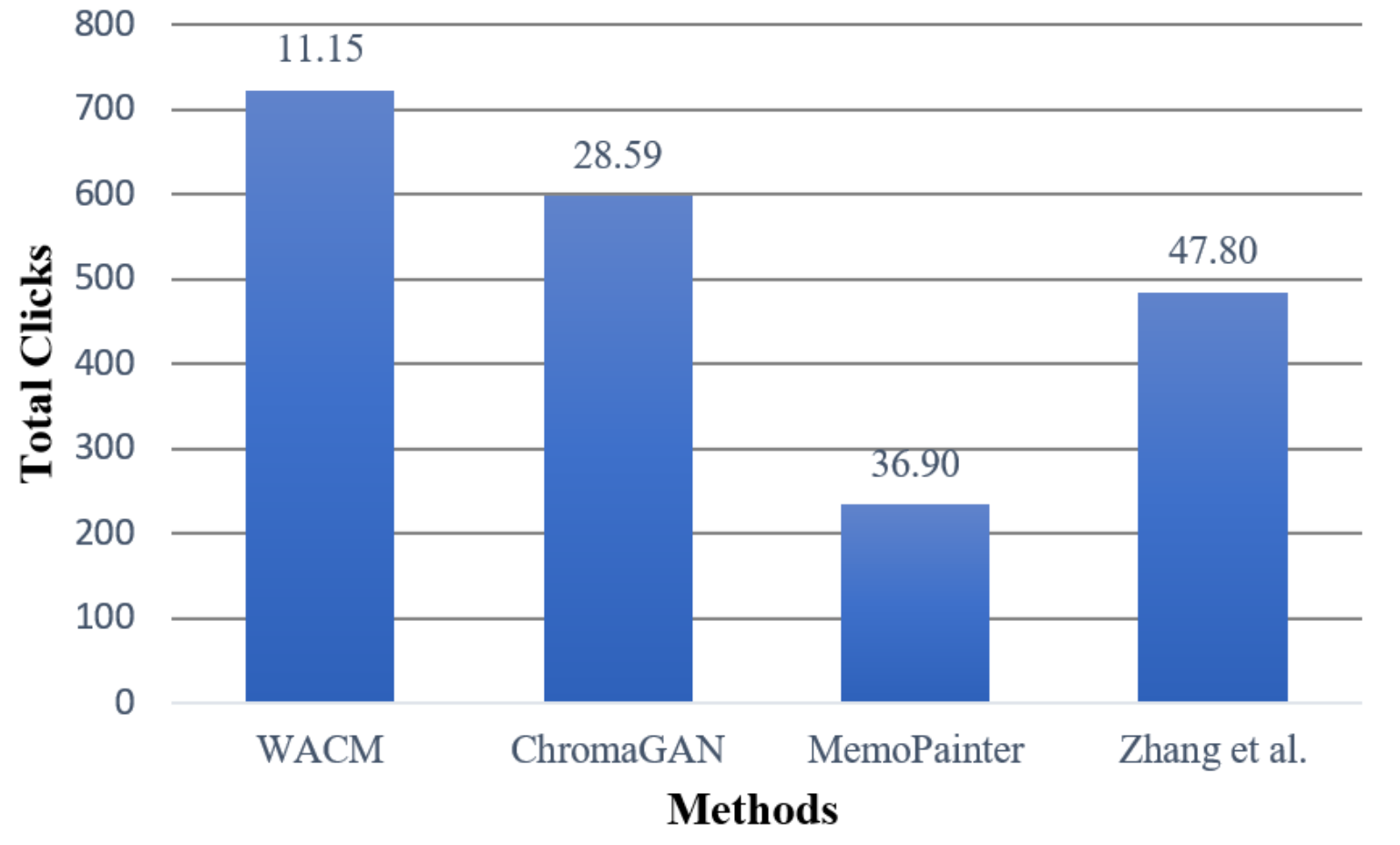}
  \caption{\hspace{-0.4em}{The total value and standard deviation (shown above the bar) of user clicks for four colorization results obtained by different methods. 
}}\label{clicks}
  \end{center}
  \vspace{-0.5cm}
\end{figure}

During the experiment, the users are asked to choose one of each pair that looks more natural. The total number of user preferences (clicks) for each colorization result is recorded, which is shown in Fig. \ref{clicks}. The highest clicks imply that the results of WACM is mostly preferred by users. Besides, the lowest standard deviation indicates that colorization results of WACM are always satisfactory despite in different image content.

\begin{table}[]
\small
\renewcommand{\arraystretch}{1.2}
    \centering
    \caption{Naturalness measurement results of LOE.}
    \vspace{-0.1cm}
    \begin{tabular}{c|c|c|c}
        \hline\hline
        \textbf{Algorithm} & \textbf{LSUN-church} & \textbf{LSUN-bedroom} & \textbf{COCO-stuff} \\ 
        \hline
        Zhang \emph{et al.} & 554,55 & {1221.14}& {839.53} \\
        \hline
        MemoPainter & {634.21} & 4590.30  & {4573.79}\\
        \hline
        {ChromaGAN} & {568.55} &{4609.66} & {4600.94}\\
        \hline
        {iGM-6C} & {\textbf{482.75}} & {1013.50} & \textbf{470.11} \\
        \hline
        {WACM} & {551.14} &{ \textbf{994.86}} & {791.35} \\
        \hline\hline 
    \end{tabular}
    \label{LOE}
\end{table}

\subsection{Ablation Study} Three main components are critical to the performance of the final WACM: prior learning in wavelet domain, structure-consistency that enforced in wavelet domain, and training high-frequency and low-frequency wavelet coefficients jointly. Here several ablation studies are conducted to validate these important designs.

\subsubsection{\textbf{Prior Learning in Wavelet or Intensity Domain}}
\begin{figure}[]
  \begin{center}
  \includegraphics[width=3.5in]{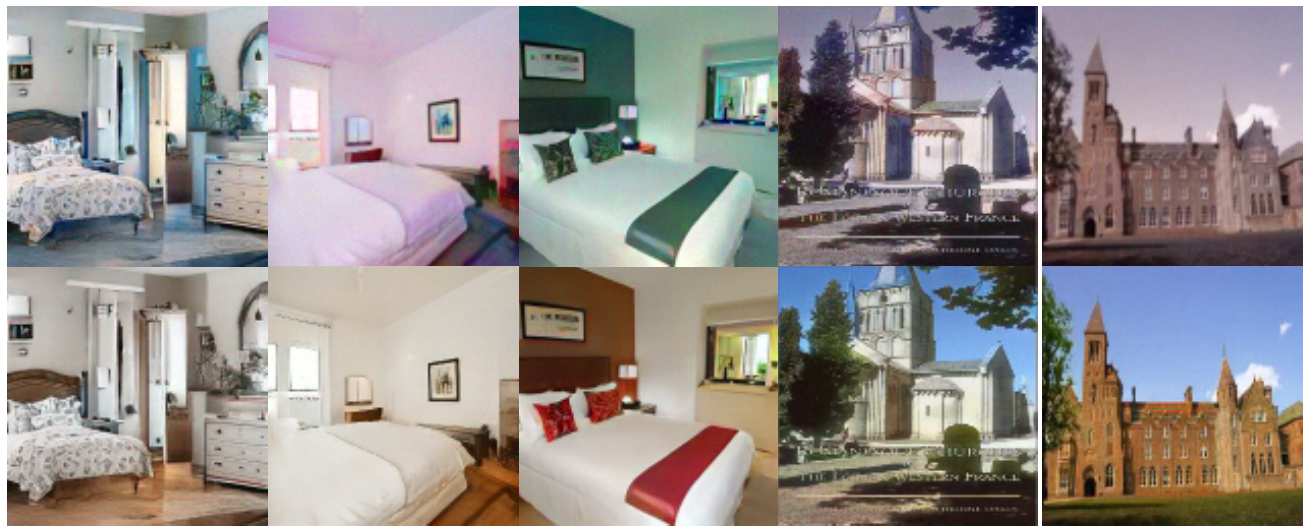}
  \caption{\hspace{-0.6em}{Visual comparisons between prior learning in 
intensity domain (the first line) and wavelet domain (the second line).
}}\label{prelearn}
  \end{center}
  \vspace{-0.5cm}
\end{figure}

We conduct an experiment to quantify the key factor of this research—training DSM in wavelet domain. We report the quantitative comparisons of prior learning in wavelet domain and intensity domain on LSUN-church and LSUN-bedroom datasets in Table \ref{ablation_domain} and exhibit some examples in Fig. \ref{prelearn}. The results present a significant performance boost gained by our method in all metrics, which further highlights the contribution of prior learning in wavelet domain. The significant improvement of SSIM is worth noting. For example, it increases by 0.07 in LSUN-church dataset, and 0.08 in LSUN-bedroom dataset, which is benefited from the complete description of details and texture of the image at all available scales via DWT.

\subsubsection{\textbf{Iteration with Different Consistencies}} 
\begin{figure*}[]
  \begin{center}
  \includegraphics[width=6.7in]{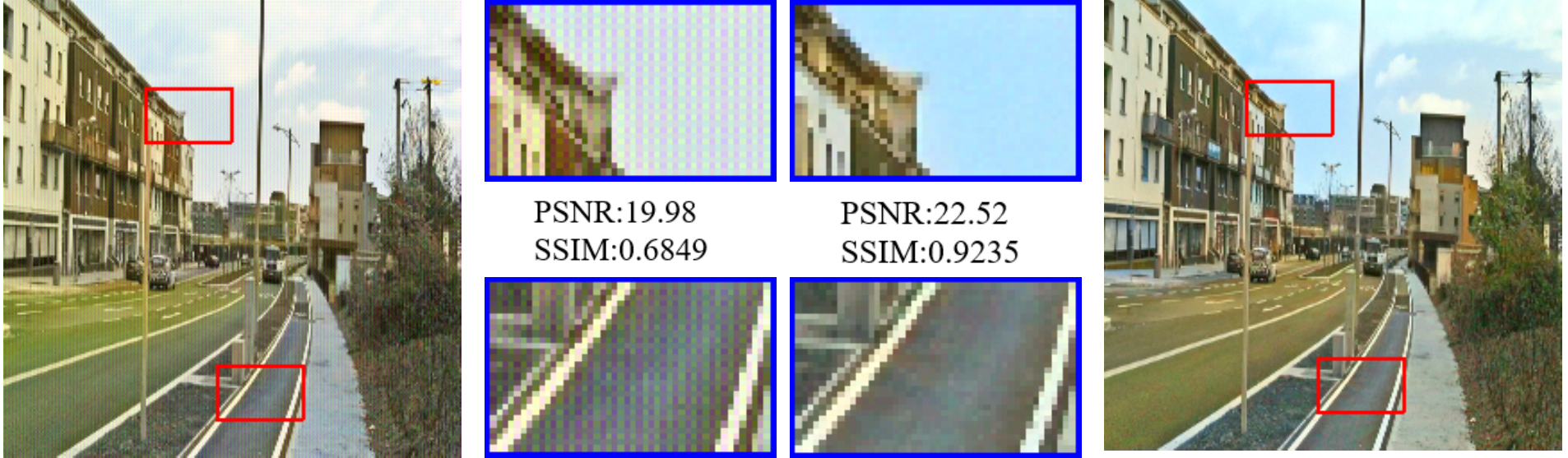}
  \leftline{ \qquad \qquad (a) WACM without SC  \hspace{8cm}  \qquad        (b) WACM with SC}
  \vspace{-0.5cm}
  \caption{\hspace{-0.7em}{Quantitative and qualitative comparisons between {(a) WACM without SC and (b) WACM with both DC and SC.} The comparison results in (a)(b) validate a performance boost gained by using SC. It can be noticed that both the PSNR and SSIM values improved under the constraint of SC as well as eliminating the “gridding” artifacts visually, which demonstrates that SC helps to achieve finer results.
}}\label{ablation}
  \end{center}
  \vspace{-0.5cm}
\end{figure*}

This ablation study is conducted to investigate the contribution of SC in wavelet domain. WACM is sampling under two different cases: without/with SC term. Fig. \ref{ablation} provides qualitative and quantitative comparisons. As shown, although the model without SC accomplishes the correct colorization with high saturation overall, the results suffer from improper gridding effects in details, which cause lower PSNR and SSIM values and imperfect visual effects.

\begin{table}[]
\small
\renewcommand{\arraystretch}{1.2}
    \centering
    \caption{Quantitative comparisons for the ablation study of prior learning in different domains on LUSN-church and LSUN-bedroom datasets.}
    \begin{tabular}{p{72 pt}|p{64pt}|p{70pt}}
         \hline\hline
         \makecell[c]{\textbf{Domain}} & \makecell[c]{\textbf{LSUN-church}} &  \makecell[c]{\textbf{LSUN-bedroom}}  \\
         \hline\hline
         \makecell[c]{Intensity domain}  & \makecell[c]{22.67/0.8584} & \makecell[c]{20.29/0.8150} \\
         \hline
         \makecell[c]{Wavelet domain} & \makecell[c]{\textbf{25.44}/\textbf{0.9265} }& \makecell[c]{\textbf{22.92}/\textbf{0.8975}}  \\
         \hline\hline
    \end{tabular}
    \label{ablation_domain}
\end{table}

However, the model with SC can constrain the generation of high-frequency wavelet coefficients and guide them toward the correct distribution, thus effectively eliminating the gridding artifacts. We also zoom in on the partial map for observation. It can be observed that the result in Fig. \ref{ablation}(b) retains the benefits of high saturation and proper color while reducing the improper effects appearing in Fig. \ref{ablation}(a). 
The results demonstrate the effectiveness using SC, especially in terms of SSIM value, which is a metric for structural characteristics. This experiment demonstrates that SC operation indeed helps to achieve finer results.

\subsubsection{\textbf{Training Wavelet Coefficients Jointly or Separately}} 
\begin{figure}[]
  \begin{center}
  \includegraphics[width=3.4in]{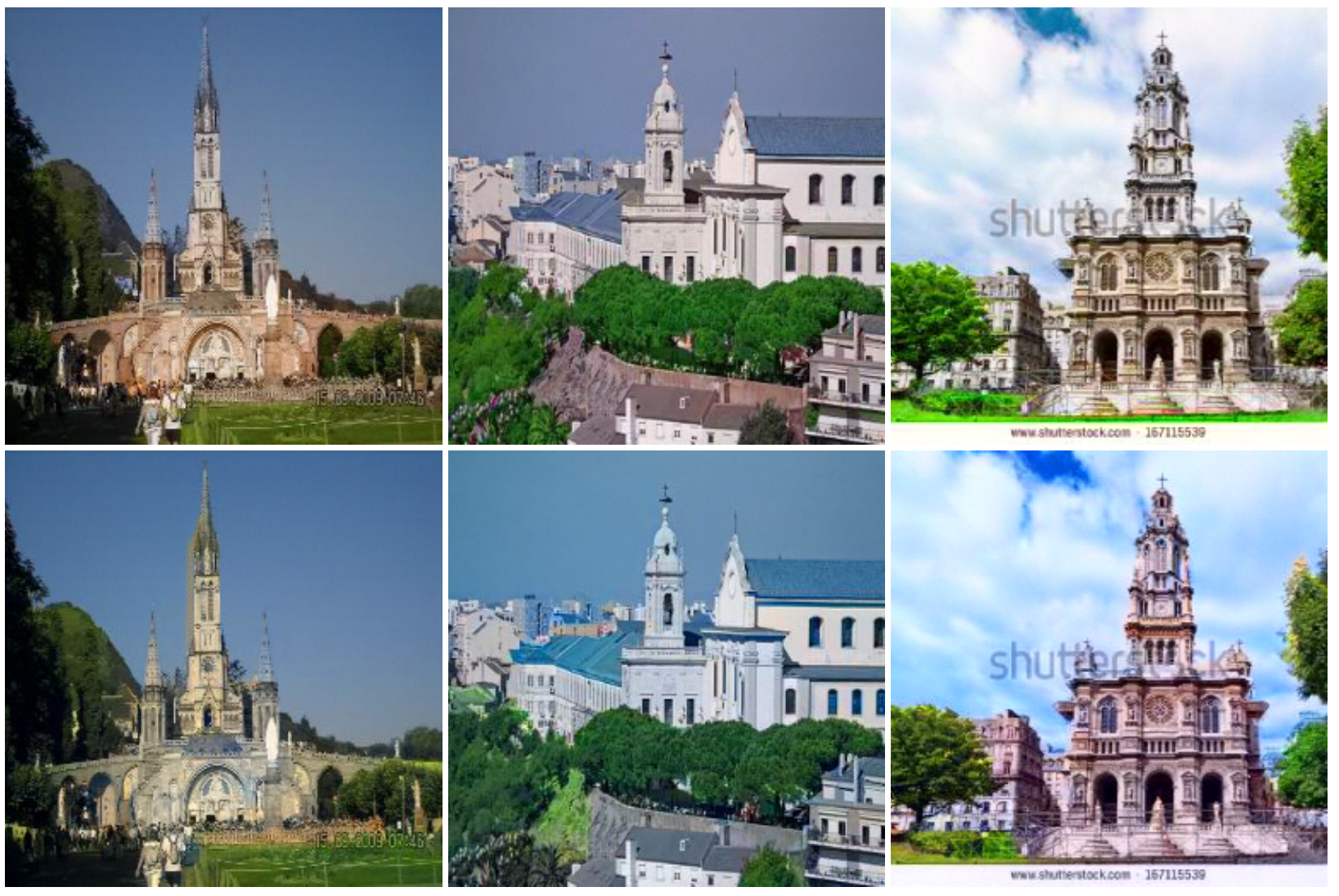}
  \vspace{-0.1cm}
  \caption{\hspace{-0.6em}{Colorization comparison of the proposed WACM between separate training and joint training of high-frequency and low-frequency wavelet coefficients. Top line: WACM-joint, Bottom line: WACM-divide.
}}\label{ablation_jors_img}
  \end{center}
  \vspace{-0.5cm}
\end{figure}

In this experiment, we investigate the colorization performance {in} two settings: joint training or separate training of high-frequency and low-frequency wavelet coefficients, namely WACM-joint and WACM-divide. The quantitative comparisons are conducted on LSUN-church dataset to evaluate their performance. Fig. \ref{ablation_jors_img} and Table \ref{ablation_jord} list the comparison results. Generally, thanks to prior learning and sampling in wavelet domain, both of them can produce satisfactory results. However, results in Table \ref{ablation_jord} present a performance boost gained by WACM-joint.

An important reason for the superior performance of joint training is that training separately cannot guarantee the consistency of the high-frequency and low-frequency wavelet coefficients generated by the network. Another reason {is that} sampling in the multi-channel embedding space is more effective than the information obtained from original objects \cite{liu2020highly}. In addition, a single network can effectively reduce the computation cost{, which} improves the efficiency of the model.

\subsection{Robustness Test}
Due to the wide application of colorization task but the datasets in real-world may {be} insufficient, it is impossible to train the model with all types of images. Therefore, the robustness of model, i.e., one model for tackling various images in different datasets, is necessary. Considering natural images contain the multiple types of potential priors, in this section, we use a model only trained by COCO-stuff to handle a variety of colorization tasks, including legacy black-and-white photos and cartoons.

\begin{table}[]
\small
\renewcommand{\arraystretch}{1.2}
    \centering
    \caption{Colorization comparison of WACM-joint and WACM-divide in the 128×128 image of LSUN-church dataset. }
    \begin{tabular}{p{72 pt}|p{64pt}|p{70pt}}
         \hline\hline
         \makecell[c]{\textbf{Algorithm} }& \makecell[c]{\textbf{PSNR} }&  \makecell[c]{\textbf{SSIM}} \\
         \hline\hline
         \makecell[c]{WACM-joint } & \makecell[c]{\textbf{25.44}}  & \makecell[c]{\textbf{0.9265} }\\
         \hline
         \makecell[c]{WACM-divide} & \makecell[c]{22.71 } & \makecell[c]{0.9023} \\
         \hline\hline
    \end{tabular}
    \label{ablation_jord}
    \vspace{-0.2cm}
\end{table}

\subsubsection{\textbf{Colorizing Legacy Black-and-White Photos}} Different from colorizing the pictures from the test datasets, which processes the original color images to obtain the grayscale images and then colorize them. In more general cases, we can only observe the grayscale image ${y}$ without knowing its forward model ${F}$. In this circumstance, the task of “blind” colorization is more challenging.

In this experiment, a prevailing processing method of forming ${F}$ is chosen:
\begin{equation}\begin{split}\label{DC}\begin{aligned}
    F(x)=(x_R+x_G+x_B)/3.0.
 \end{aligned}
 \end{split}
\end{equation}

\begin{figure}[t]
  \begin{center}
  \includegraphics[width=3.4in]{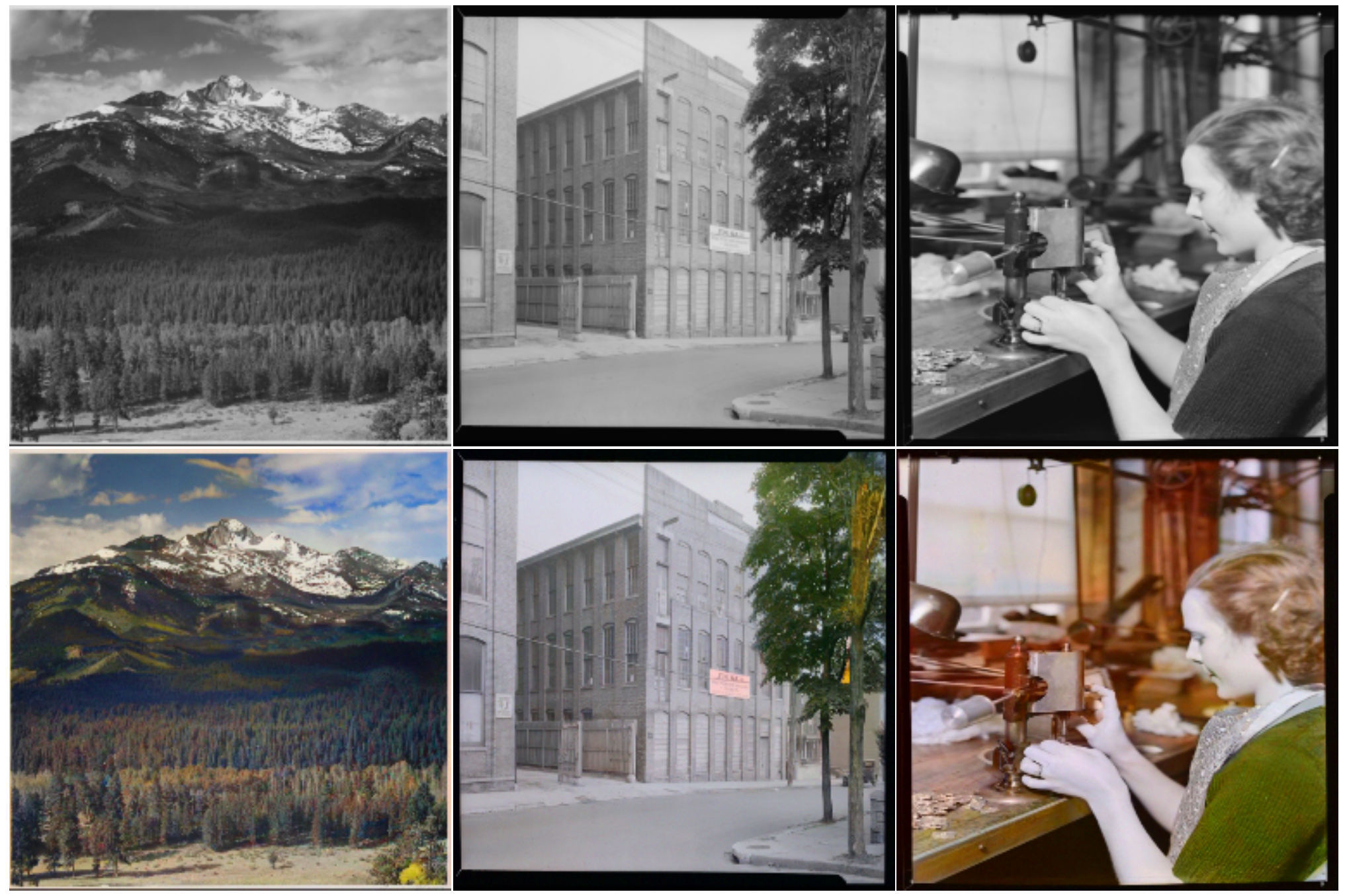}
  \caption{\hspace{-0.5em}{Colorizing legacy black and white photographs. Our model can obtain realistic colorization results whether it is a picture of landscapes or close-ups. The images we choose are (a) Colorado National Park, 1941 (b) Textile Mill, June 1937 (c) Hamilton, 1936.
}}\label{black}
  \end{center}
  \vspace{-0.5cm}
\end{figure}

As observed in Fig. \ref{black}, convincing results are generated by WACM. Taking the first picture for example, the results are realistic in terms of texture, contrast and saturation.

\subsubsection{\textbf{Colorizing Cartoons}}  When it comes to real-world applications, cartoons and animation are two main areas needed for colorization. However, data for animations and cartoons {is} often limited as the cartoon images are difficult to create and must be colored by hand. This problem can be alleviated by training the model {on} natural image datasets that have abundant images and then applying it to cartoon colorization.

\begin{figure}[t]
  \begin{center}
  \includegraphics[width=3.5in]{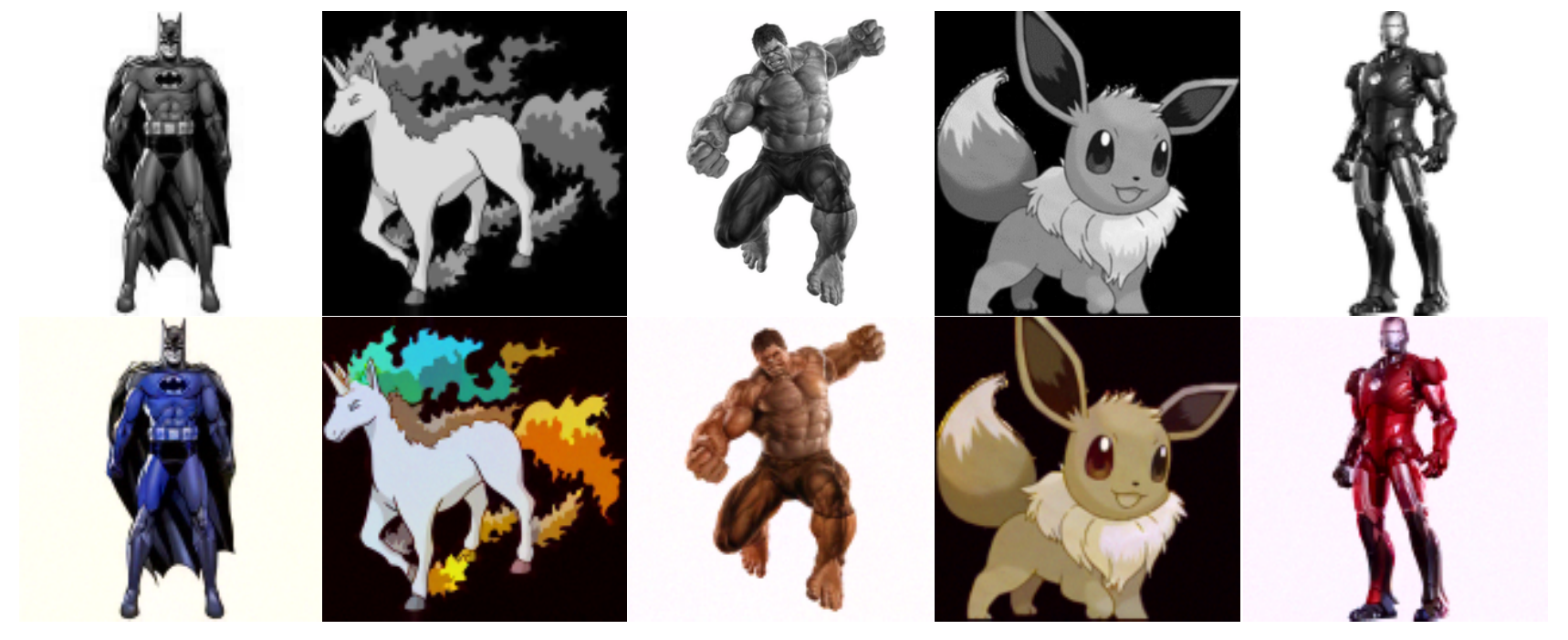}
  \caption{\hspace{-0.3em}{Colorize cartoon images. The top line involves the observed grayscale image. The bottom line lists the results obtained by WACM. Notice that all these results are obtained automatically by WACM without model re-training in the cartoon dataset.
}}\label{cartoon}
  \end{center}
  \vspace{-0.5cm}
\end{figure}

In this experiment, we try to learn wavelet prior from the COCO-stuff dataset and apply it to colorize cartoons. {Some} results of WACM are exhibited in Fig. \ref{cartoon}. Although the accuracy of manual colorization cannot be achieved, the results produced by WACM are satisfactory and quite good. As can be seen in the second image, the cartoon characters are colored in blue and orange and {have} obtained color consistency. In the fifth image, the textures of the character (the metallic texture of the character's body) are retained as well.

In most cases, WACM can produce realistic and satisfactory results. Notably, in these experiments, WACM is only trained on the COCO-stuff dataset. This phenomenon indicates the effectiveness and robustness of WACM.

\subsection{Colorization Diversity} Image colorization is essentially a one-to-many task as multiple feasible colorized results can be given for the same grayscale input. Generating a diverse set of colorization solutions is an effective way to tackle this multi-modality challenge. In general, it can be achieved via generative models.

Leveraging the generative model as well as multi-scale and multi-channel  representation in wavelet domain, our model can generate multiple feasible colorized images to meet different needs. Some diverse colorization results are shown in Fig. 16. The results show that our generated colored images have fine-grained and vibrant colors.

\begin{figure}[t]
  \begin{center}
  \includegraphics[width=3.4in]{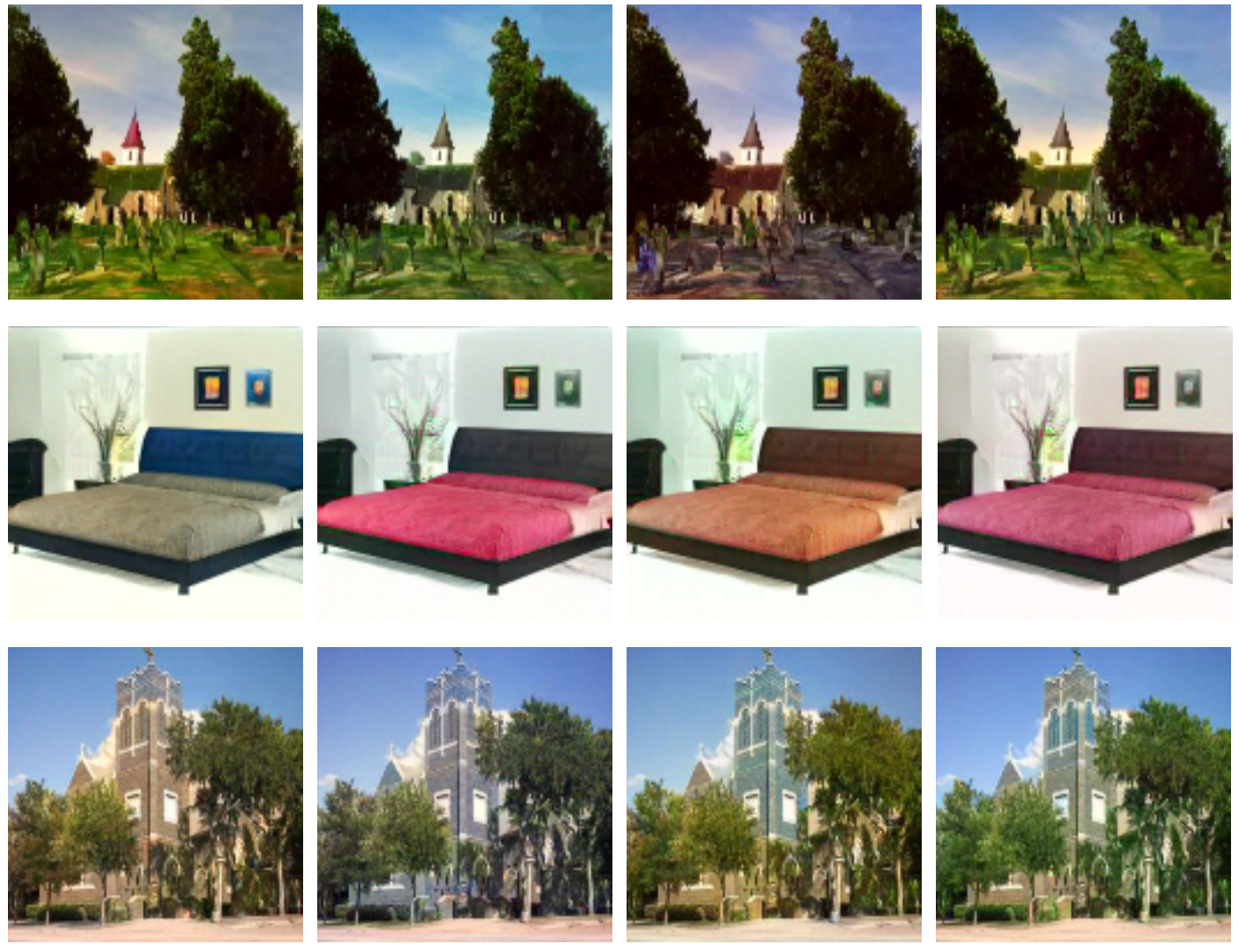}
  \caption{\hspace{-0.6em}{Illustration of the diverse colorization by WACM. For each image, WACM produces twelve colorized samples, from which four different styles are chosen. It can be noticed that WACM can produce various styles for a single image.}}\label{diversity}
  \end{center}
  \vspace{-0.5cm}
\end{figure}

\begin{figure}[t]
  \begin{center}
  \includegraphics[width=3.4in]{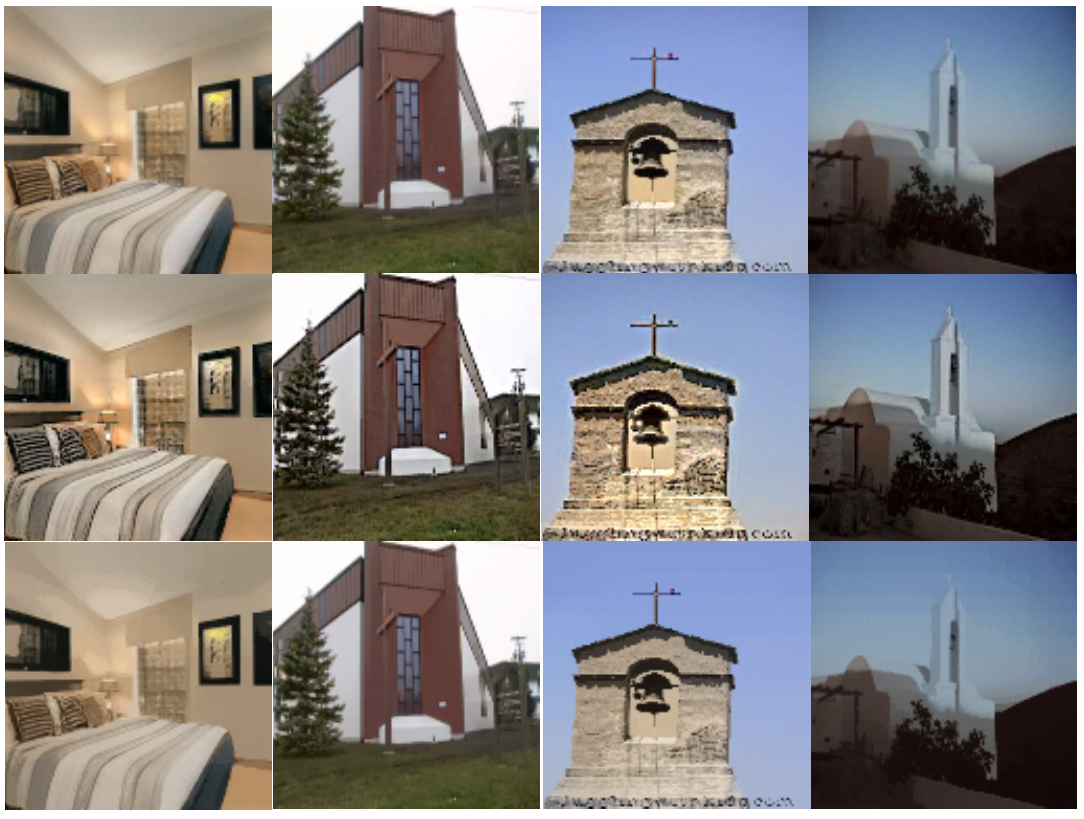}
  \caption{\hspace{-0.6em}{Colorization comparison of the original results and those produced after pre-processing or post-processing techniques. Top line: WACM without image enhancement. Second line: WACM with image enhancement. Bottom line: WACM with filtering.
}}\label{enhance}
  \end{center}
  \vspace{-0.5cm}
\end{figure}

\subsection{{Pre-processing and Post-processing}} {To further improve the subjective quality and reduce visual artifacts of colorized images, some pre-processing and post-processing techniques via image enhancement and filtering can be used to help generate more pleasing results.}
{\subsubsection{\textbf{Pre-processing via Image Enhancement}} We can pre-process grayscale images using image enhancement techniques before feeding them into the colorization network. Motivated by \cite{ghosh2019saliency}, in this experiment, we pre-process the grayscale image by enhancing the details of salient areas. To be more specific, we employ frequency-tuned methods\cite{achanta2009frequency} to detect the salient regions, then the details in these regions are enhanced by the multi-scale detail boosting method\cite{kim2015dark}.}
\subsubsection{\textbf{Post-processing via Image Filtering}} Considering that colorization is a highly free task, the generated colorization results may have problems such as existing noise or color overflow. Thus, some post-processing such as Gaussian filtering\cite{shin2005block} or bilateral filtering\cite{ghosh2016fast,ghosh2018color} can be applied to improve the quality of the colorization images. We initially used method \cite{barron2016fast} for this post-processing experiment. As can be seen from Fig. \ref{enhance}, by pre-processing the images via image enhancement, the colorization results become sharper, and the edges of objects are clearer. In addition, after post-processing via filtering, the noise in the images is suppressed, and the colorization results become smoother. 
Therefore, by combining the model with pre-processing or post-processing techniques, we can obtain different styles of results with higher subjective quality.
\subsection{Limitations in Real-world Applications}
Extensive experiments have demonstrated that WACM is capable of producing diverse and high-quality colorization results. When it comes to real-world applications, there are still two existing limitations under some circumstances: 1) Colorization degradation caused by data distribution; 2) Computational burden.

First, different from approximating the potential mapping from grayscale to color directly, WACM tackles colorization task using MCMC method. In most cases, WACM has satisfactory robustness and generalization thanks to the proposed strategy, while performance degradation still occasionally occurs when dealing with some images that has different categories or content to the training set, especially when their colored version happens to locate in the low-density regions of the learned distribution. In practical applications, a possible solution is to further process training set and training schedule according to the specific application scenario to learn a more appropriate probability density distribution.

Second, since the colorization process of WACM is conducted in iterative manner, and a larger number of iterations at each noise level generally contributes to better samples. Thus, it achieves better performance at the price of more computation compared with some supervised learning methods in end-to-end manner. This limitation could be alleviated by parallelizing sampling process, or striking a suitable trade-off between performance and computation by choosing noise levels.

\section{Conclusions}
To summarize, this work proposed an iterative generative model in wavelet domain to address the colorization problem. We have shown that utilizing the multi-scale and multi-channel {strategy} to make the prior learning procedure in {a} lower-dimensional {and more statistically informative} subspace via wavelet transform is an effective optimization scheme to improve the performance of score-based generative models. By taking advantage of the non-redundant and multi-scale representation of DWT and the high-precision reconstruction of IDWT, we address some general problems in {generative modeling}. Meanwhile, two consistency terms are proposed to make full use of wavelet transform in colorization while avoiding the improper effects caused by the uncertainty of generative model. Extensive experiments have been conducted to demonstrate that the proposed method achieves state-of-the-art performance in automatic colorization, and shows strong superiority over the previous methods in both quantitative and qualitative assessments.

{For future work, research could continue to explore and apply the proposed method to few-shot image colorization task. Combining it with meta-learning\cite{nichol2018first} and clustering methods such as spectral clustering\cite{li2018dynamic,li2018rank} to construct a model that can produce excellent colorization results even without a large quantity of data. Also, the effectiveness of multi-level wavelet transform prior can be further explored, as well as the most efficient way to utilize it in score-based generative model. Furthermore, the hyperparameter optimization problem using modern statistical methods such as Bayesian optimization\cite{kaselimi2019bayesian,feurer2019hyperparameter} for both the estimation and sampling process of the score-based generative model in wavelet domain is also a direction worthy of more exploration in the future.
}

\ifCLASSOPTIONcaptionsoff,，
  \newpage
\fi




\bibliographystyle{IEEEtran}
\bibliography{IEEEabrv}

\begin{thebibliography}{10}
\providecommand{\url}[1]{#1}
\csname url@rmstyle\endcsname
\providecommand{\newblock}{\relax}
\providecommand{\bibinfo}[2]{#2}
\providecommand\BIBentrySTDinterwordspacing{\spaceskip=0pt\relax}
\providecommand\BIBentryALTinterwordstretchfactor{4}
\providecommand\BIBentryALTinterwordspacing{\spaceskip=\fontdimen2\font plus
\BIBentryALTinterwordstretchfactor\fontdimen3\font minus
  \fontdimen4\font\relax}
\providecommand\BIBforeignlanguage[2]{{%
\expandafter\ifx\csname l@#1\endcsname\relax
\typeout{** WARNING: IEEEtran.bst: No hyphenation pattern has been}%
\typeout{** loaded for the language `#1'. Using the pattern for}%
\typeout{** the default language instead.}%
\else
\language=\csname l@#1\endcsname
\fi
#2}}
\renewcommand\BIBentryALTinterwordstretchfactor{4}

\bibitem{fatima2021grey}
A.~Fatima, W.~Hussain, and S.~Rasool, ``Grey is the new rgb: How good is
  gan-based image colorization for image compression?'' \emph{Multimedia Tools
  and Applications}, vol.~80, no.~3, pp. 3775--3791, 2021.

\bibitem{baig2017multiple}
M.~H. Baig and L.~Torresani, ``Multiple hypothesis colorization and its
  application to image compression,'' \emph{Computer Vision and Image
  Understanding}, vol. 164, pp. 111--123, 2017.

\bibitem{bian2021deep}
Y.~Bian, Y.~Jiang, Y.~Huang, X.~Yang, W.~Deng, H.~Shen, R.~Shen, and C.~Kuang,
  ``Deep learning virtual colorization overcoming chromatic aberrations in
  singlet lens microscopy,'' \emph{APL Photonics}, vol.~6, no.~3, p. 031301,
  2021.

\bibitem{levin2004colorization}
A.~Levin, D.~Lischinski, and Y.~Weiss, ``Colorization using optimization,'' in
  \emph{ACM SIGGRAPH 2004 Papers}, 2004, pp. 689--694.

\bibitem{huang2005adaptive}
Y.-C. Huang, Y.-S. Tung, J.-C. Chen, S.-W. Wang, and J.-L. Wu, ``An adaptive
  edge detection based colorization algorithm and its applications,'' in
  \emph{Proceedings of the 13th annual ACM international conference on
  Multimedia}, 2005, pp. 351--354.

\bibitem{qu2006manga}
Y.~Qu, T.-T. Wong, and P.-A. Heng, ``Manga colorization,'' \emph{ACM
  Transactions on Graphics (TOG)}, vol.~25, no.~3, pp. 1214--1220, 2006.

\bibitem{luan2007natural}
Q.~Luan, F.~Wen, D.~Cohen-Or, L.~Liang, Y.-Q. Xu, and H.-Y. Shum, ``Natural
  image colorization,'' in \emph{Proceedings of the 18th Eurographics
  conference on Rendering Techniques}, 2007, pp. 309--320.

\bibitem{welsh2002transferring}
T.~Welsh, M.~Ashikhmin, and K.~Mueller, ``Transferring color to greyscale
  images,'' in \emph{Proceedings of the 29th annual conference on Computer
  graphics and interactive techniques}, 2002, pp. 277--280.

\bibitem{ironi2005colorization}
R.~Ironi, D.~Cohen-Or, and D.~Lischinski, ``Colorization by example.''
  \emph{Rendering techniques}, vol.~29, pp. 201--210, 2005.

\bibitem{charpiat2008automatic}
G.~Charpiat, M.~Hofmann, and B.~Sch{\"o}lkopf, ``Automatic image colorization
  via multimodal predictions,'' in \emph{European conference on computer
  vision}.\hskip 1em plus 0.5em minus 0.4em\relax Springer, 2008, pp. 126--139.

\bibitem{chia2011semantic}
A.~Y.-S. Chia, S.~Zhuo, R.~K. Gupta, Y.-W. Tai, S.-Y. Cho, P.~Tan, and S.~Lin,
  ``Semantic colorization with internet images,'' \emph{ACM Transactions on
  Graphics (TOG)}, vol.~30, no.~6, pp. 1--8, 2011.

\bibitem{deshpande2015learning}
A.~Deshpande, J.~Rock, and D.~Forsyth, ``Learning large-scale automatic image
  colorization,'' in \emph{Proceedings of the IEEE International Conference on
  Computer Vision}, 2015, pp. 567--575.

\bibitem{yoo2019coloring}
S.~Yoo, H.~Bahng, S.~Chung, J.~Lee, J.~Chang, and J.~Choo, ``Coloring with
  limited data: Few-shot colorization via memory augmented networks,'' in
  \emph{Proceedings of the IEEE/CVF Conference on Computer Vision and Pattern
  Recognition}, 2019, pp. 11\,283--11\,292.

\bibitem{suarez2017infrared}
P.~L. Su{\'a}rez, A.~D. Sappa, and B.~X. Vintimilla, ``Infrared image
  colorization based on a triplet dcgan architecture,'' in \emph{Proceedings of
  the IEEE Conference on Computer Vision and Pattern Recognition Workshops},
  2017, pp. 18--23.

\bibitem{vitoria2020chromagan}
P.~Vitoria, L.~Raad, and C.~Ballester, ``Chromagan: Adversarial picture
  colorization with semantic class distribution,'' in \emph{Proceedings of the
  IEEE/CVF Winter Conference on Applications of Computer Vision}, 2020, pp.
  2445--2454.

\bibitem{zhou2020progressive}
J.~Zhou, K.~Hong, T.~Deng, Y.~Wang, and Q.~Liu, ``Progressive colorization via
  iterative generative models,'' \emph{IEEE Signal Processing Letters},
  vol.~27, pp. 2054--2058, 2020.

\bibitem{cao2017unsupervised}
Y.~Cao, Z.~Zhou, W.~Zhang, and Y.~Yu, ``Unsupervised diverse colorization via
  generative adversarial networks,'' in \emph{Joint European conference on
  machine learning and knowledge discovery in databases}.\hskip 1em plus 0.5em
  minus 0.4em\relax Springer, 2017, pp. 151--166.

\bibitem{zhang2016colorful}
R.~Zhang, P.~Isola, and A.~A. Efros, ``Colorful image colorization,'' in
  \emph{European conference on computer vision}.\hskip 1em plus 0.5em minus
  0.4em\relax Springer, 2016, pp. 649--666.

\bibitem{iizuka2016let}
S.~Iizuka, E.~Simo-Serra, and H.~Ishikawa, ``Let there be color! joint
  end-to-end learning of global and local image priors for automatic image
  colorization with simultaneous classification,'' \emph{ACM Transactions on
  Graphics (ToG)}, vol.~35, no.~4, pp. 1--11, 2016.

\bibitem{isola2017image}
P.~Isola, J.-Y. Zhu, T.~Zhou, and A.~A. Efros, ``Image-to-image translation
  with conditional adversarial networks,'' in \emph{Proceedings of the IEEE
  conference on computer vision and pattern recognition}, 2017, pp. 1125--1134.

\bibitem{zhao2020pixelated}
J.~Zhao, J.~Han, L.~Shao, and C.~G. Snoek, ``Pixelated semantic colorization,''
  \emph{International Journal of Computer Vision}, vol. 128, no.~4, pp.
  818--834, 2020.

\bibitem{deshpande2017learning}
A.~Deshpande, J.~Lu, M.-C. Yeh, M.~Jin~Chong, and D.~Forsyth, ``Learning
  diverse image colorization,'' in \emph{Proceedings of the IEEE Conference on
  Computer Vision and Pattern Recognition}, 2017, pp. 6837--6845.

\bibitem{vincent2008extracting}
P.~Vincent, H.~Larochelle, Y.~Bengio, and P.-A. Manzagol, ``Extracting and
  composing robust features with denoising autoencoders,'' in \emph{Proceedings
  of the 25th international conference on Machine learning}, 2008, pp.
  1096--1103.

\bibitem{vincent2011connection}
P.~Vincent, ``A connection between score matching and denoising autoencoders,''
  \emph{Neural computation}, vol.~23, no.~7, pp. 1661--1674, 2011.

\bibitem{jayaram2020source}
V.~Jayaram and J.~Thickstun, ``Source separation with deep generative priors,''
  in \emph{International Conference on Machine Learning}.\hskip 1em plus 0.5em
  minus 0.4em\relax PMLR, 2020, pp. 4724--4735.

\bibitem{song2019generative}
Y.~Song and S.~Ermon, ``Generative modeling by estimating gradients of the data
  distribution,'' \emph{arXiv preprint arXiv:1907.05600}, 2019.

\bibitem{narayanan2010sample}
H.~Narayanan and S.~Mitter, ``Sample complexity of testing the manifold
  hypothesis,'' in \emph{Proceedings of the 23rd International Conference on
  Neural Information Processing Systems-Volume 2}, 2010, pp. 1786--1794.

\bibitem{rifai2011manifold}
S.~Rifai, Y.~N. Dauphin, P.~Vincent, Y.~Bengio, and X.~Muller, ``The manifold
  tangent classifier,'' \emph{Advances in neural information processing
  systems}, vol.~24, pp. 2294--2302, 2011.

\bibitem{quan2021homotopic}
C.~Quan, J.~Zhou, Y.~Zhu, Y.~Chen, S.~Wang, D.~Liang, and Q.~Liu, ``Homotopic
  gradients of generative density priors for mr image reconstruction,''
  \emph{IEEE Transactions on Medical Imaging}, 2021.

\bibitem{zhou2021learning}
Y.~Zhou, C.~Chen, and J.~Xu, ``Learning high-dimensional distributions with
  latent neural fokker-planck kernels,'' \emph{arXiv preprint
  arXiv:2105.04538}, 2021.

\bibitem{block2020fast}
A.~Block, Y.~Mroueh, A.~Rakhlin, and J.~Ross, ``Fast mixing of multi-scale
  langevin dynamics underthe manifold hypothesis,'' \emph{arXiv preprint
  arXiv:2006.11166}, 2020.

\bibitem{akansu2001multiresolution}
A.~N. Akansu, R.~A. Haddad, P.~A. Haddad, and P.~R. Haddad,
  \emph{Multiresolution signal decomposition: transforms, subbands, and
  wavelets}.\hskip 1em plus 0.5em minus 0.4em\relax Academic press, 2001.

\bibitem{zhang2019wavelet}
D.~Zhang, ``Wavelet transform,'' in \emph{Fundamentals of Image Data
  Mining}.\hskip 1em plus 0.5em minus 0.4em\relax Springer, 2019, pp. 35--44.

\bibitem{acharya2020image}
M.~Acharya, S.~Poddar, A.~Chakrabarti, and H.~Rahaman, ``Image classification
  based on approximate wavelet transform and transfer learning on deep
  convolutional neural networks,'' in \emph{2020 International Symposium on
  Devices, Circuits and Systems (ISDCS)}.\hskip 1em plus 0.5em minus
  0.4em\relax IEEE, 2020, pp. 1--6.

\bibitem{guo2017deep}
T.~Guo, H.~Seyed~Mousavi, T.~Huu~Vu, and V.~Monga, ``Deep wavelet prediction
  for image super-resolution,'' in \emph{Proceedings of the IEEE Conference on
  Computer Vision and Pattern Recognition Workshops}, 2017, pp. 104--113.

\bibitem{sharma2016satellite}
A.~Sharma and A.~Khunteta, ``Satellite image contrast and resolution
  enhancement using discrete wavelet transform and singular value
  decomposition,'' in \emph{2016 International Conference on Emerging Trends in
  Electrical Electronics \& Sustainable Energy Systems (ICETEESES)}.\hskip 1em
  plus 0.5em minus 0.4em\relax IEEE, 2016, pp. 374--378.

\bibitem{ghazali2007feature}
K.~H. Ghazali, M.~F. Mansor, M.~M. Mustafa, and A.~Hussain, ``Feature
  extraction technique using discrete wavelet transform for image
  classification,'' in \emph{2007 5th Student Conference on Research and
  Development}.\hskip 1em plus 0.5em minus 0.4em\relax IEEE, 2007, pp. 1--4.

\bibitem{zhu1998study}
C.~Zhu and X.~Yang, ``Study of remote sensing image texture analysis and
  classification using wavelet,'' \emph{International Journal of Remote
  Sensing}, vol.~19, no.~16, pp. 3197--3203, 1998.

\bibitem{chowdhury2012image}
M.~M.~H. Chowdhury and A.~Khatun, ``Image compression using discrete wavelet
  transform,'' \emph{International Journal of Computer Science Issues (IJCSI)},
  vol.~9, no.~4, p. 327, 2012.

\bibitem{liu2020highly}
Q.~Liu, Q.~Yang, H.~Cheng, S.~Wang, M.~Zhang, and D.~Liang, ``Highly
  undersampled magnetic resonance imaging reconstruction using autoencoding
  priors,'' \emph{Magnetic resonance in medicine}, vol.~83, no.~1, pp.
  322--336, 2020.

\bibitem{anwar2020image}
S.~Anwar, M.~Tahir, C.~Li, A.~Mian, F.~S. Khan, and A.~W. Muzaffar, ``Image
  colorization: A survey and dataset,'' \emph{arXiv preprint arXiv:2008.10774},
  2020.

\bibitem{stankovic2003haar}
R.~S. Stankovi{\'c} and B.~J. Falkowski, ``The haar wavelet transform: its
  status and achievements,'' \emph{Computers \& Electrical Engineering},
  vol.~29, no.~1, pp. 25--44, 2003.

\bibitem{lin2000feature}
J.~Lin and L.~Qu, ``Feature extraction based on morlet wavelet and its
  application for mechanical fault diagnosis,'' \emph{Journal of sound and
  vibration}, vol. 234, no.~1, pp. 135--148, 2000.

\bibitem{vonesch2007generalized}
C.~Vonesch, T.~Blu, and M.~Unser, ``Generalized daubechies wavelet families,''
  \emph{IEEE Transactions on Signal Processing}, vol.~55, no.~9, pp.
  4415--4429, 2007.

\bibitem{salakhutdinov2009deep}
R.~Salakhutdinov and G.~Hinton, ``Deep boltzmann machines,'' in
  \emph{Artificial intelligence and statistics}.\hskip 1em plus 0.5em minus
  0.4em\relax PMLR, 2009, pp. 448--455.

\bibitem{kingma2013auto}
D.~P. Kingma and M.~Welling, ``Auto-encoding variational bayes,'' \emph{arXiv
  preprint arXiv:1312.6114}, 2013.

\bibitem{hyvarinen2005estimation}
A.~Hyv{\"a}rinen and P.~Dayan, ``Estimation of non-normalized statistical
  models by score matching.'' \emph{Journal of Machine Learning Research},
  vol.~6, no.~4, 2005.

\bibitem{robert2004monte}
C.~P. Robert, G.~Casella, and G.~Casella, \emph{Monte Carlo statistical
  methods}.\hskip 1em plus 0.5em minus 0.4em\relax Springer, 2004, vol.~2.

\bibitem{brooks2011handbook}
S.~Brooks, A.~Gelman, G.~Jones, and X.-L. Meng, \emph{Handbook of markov chain
  monte carlo}.\hskip 1em plus 0.5em minus 0.4em\relax CRC press, 2011.

\bibitem{bakry2014analysis}
D.~Bakry, I.~Gentil, M.~Ledoux, \emph{et~al.}, \emph{Analysis and geometry of
  Markov diffusion operators}.\hskip 1em plus 0.5em minus 0.4em\relax Springer,
  2014, vol. 103.

\bibitem{fefferman2016testing}
C.~Fefferman, S.~Mitter, and H.~Narayanan, ``Testing the manifold hypothesis,''
  \emph{Journal of the American Mathematical Society}, vol.~29, no.~4, pp.
  983--1049, 2016.

\bibitem{belkin2003laplacian}
M.~Belkin and P.~Niyogi, ``Laplacian eigenmaps for dimensionality reduction and
  data representation,'' \emph{Neural computation}, vol.~15, no.~6, pp.
  1373--1396, 2003.

\bibitem{sutherland2018efficient}
D.~Sutherland, H.~Strathmann, M.~Arbel, and A.~Gretton, ``Efficient and
  principled score estimation with nystr{\"o}m kernel exponential families,''
  in \emph{International Conference on Artificial Intelligence and
  Statistics}.\hskip 1em plus 0.5em minus 0.4em\relax PMLR, 2018, pp. 652--660.

\bibitem{yu2015lsun}
F.~Yu, A.~Seff, Y.~Zhang, S.~Song, T.~Funkhouser, and J.~Xiao, ``Lsun:
  Construction of a large-scale image dataset using deep learning with humans
  in the loop,'' \emph{arXiv preprint arXiv:1506.03365}, 2015.

\bibitem{caesar2018coco}
H.~Caesar, J.~Uijlings, and V.~Ferrari, ``Coco-stuff: Thing and stuff classes
  in context,'' in \emph{Proceedings of the IEEE conference on computer vision
  and pattern recognition}, 2018, pp. 1209--1218.

\bibitem{lin2014microsoft}
T.-Y. Lin, M.~Maire, S.~Belongie, J.~Hays, P.~Perona, D.~Ramanan,
  P.~Doll{\'a}r, and C.~L. Zitnick, ``Microsoft coco: Common objects in
  context,'' in \emph{European conference on computer vision}.\hskip 1em plus
  0.5em minus 0.4em\relax Springer, 2014, pp. 740--755.

\bibitem{lin2017refinenet}
G.~Lin, A.~Milan, C.~Shen, and I.~Reid, ``Refinenet: Multi-path refinement
  networks for high-resolution semantic segmentation,'' in \emph{Proceedings of
  the IEEE conference on computer vision and pattern recognition}, 2017, pp.
  1925--1934.

\bibitem{wang2013naturalness}
S.~Wang, J.~Zheng, H.-M. Hu, and B.~Li, ``Naturalness preserved enhancement
  algorithm for non-uniform illumination images,'' \emph{IEEE Transactions on
  Image Processing}, vol.~22, no.~9, pp. 3538--3548, 2013.

\bibitem{gu2020image}
J.~Gu, Y.~Shen, and B.~Zhou, ``Image processing using multi-code gan prior,''
  in \emph{Proceedings of the IEEE/CVF conference on computer vision and
  pattern recognition}, 2020, pp. 3012--3021.

\bibitem{fang2019superpixel}
F.~Fang, T.~Wang, T.~Zeng, and G.~Zhang, ``A superpixel-based variational model
  for image colorization,'' \emph{IEEE transactions on visualization and
  computer graphics}, vol.~26, no.~10, pp. 2931--2943, 2019.

\bibitem{ghosh2019saliency}
S.~Ghosh, R.~G. Gavaskar, and K.~N. Chaudhury, ``Saliency guided image detail
  enhancement,'' in \emph{2019 National Conference on Communications
  (NCC)}.\hskip 1em plus 0.5em minus 0.4em\relax IEEE, 2019, pp. 1--6.

\bibitem{achanta2009frequency}
R.~Achanta, S.~Hemami, F.~Estrada, and S.~Susstrunk, ``Frequency-tuned salient
  region detection,'' in \emph{2009 IEEE conference on computer vision and
  pattern recognition}.\hskip 1em plus 0.5em minus 0.4em\relax IEEE, 2009, pp.
  1597--1604.

\bibitem{kim2015dark}
Y.~Kim, Y.~J. Koh, C.~Lee, S.~Kim, and C.-S. Kim, ``Dark image enhancement
  based onpairwise target contrast and multi-scale detail boosting,'' in
  \emph{2015 IEEE International Conference on Image Processing (ICIP)}.\hskip
  1em plus 0.5em minus 0.4em\relax IEEE, 2015, pp. 1404--1408.

\bibitem{shin2005block}
D.-H. Shin, R.-H. Park, S.~Yang, and J.-H. Jung, ``Block-based noise estimation
  using adaptive gaussian filtering,'' \emph{IEEE Transactions on Consumer
  Electronics}, vol.~51, no.~1, pp. 218--226, 2005.

\bibitem{ghosh2016fast}
S.~Ghosh and K.~N. Chaudhury, ``Fast bilateral filtering of vector-valued
  images,'' in \emph{2016 IEEE international conference on image processing
  (ICIP)}.\hskip 1em plus 0.5em minus 0.4em\relax IEEE, 2016, pp. 1823--1827.

\bibitem{ghosh2018color}
------, ``Color bilateral filtering using stratified fourier sampling,'' in
  \emph{2018 IEEE Global Conference on Signal and Information Processing
  (GlobalSIP)}.\hskip 1em plus 0.5em minus 0.4em\relax IEEE, 2018, pp. 26--30.

\bibitem{barron2016fast}
J.~T. Barron and B.~Poole, ``The fast bilateral solver,'' in \emph{European
  Conference on Computer Vision}.\hskip 1em plus 0.5em minus 0.4em\relax
  Springer, 2016, pp. 617--632.

\bibitem{nichol2018first}
A.~Nichol, J.~Achiam, and J.~Schulman, ``On first-order meta-learning
  algorithms,'' \emph{arXiv preprint arXiv:1803.02999}, 2018.

\bibitem{li2018dynamic}
Z.~Li, F.~Nie, X.~Chang, Y.~Yang, C.~Zhang, and N.~Sebe, ``Dynamic affinity
  graph construction for spectral clustering using multiple features,''
  \emph{IEEE transactions on neural networks and learning systems}, vol.~29,
  no.~12, pp. 6323--6332, 2018.

\bibitem{li2018rank}
Z.~Li, F.~Nie, X.~Chang, L.~Nie, H.~Zhang, and Y.~Yang, ``Rank-constrained
  spectral clustering with flexible embedding,'' \emph{IEEE transactions on
  neural networks and learning systems}, vol.~29, no.~12, pp. 6073--6082, 2018.

\bibitem{kaselimi2019bayesian}
M.~Kaselimi, N.~Doulamis, A.~Doulamis, A.~Voulodimos, and E.~Protopapadakis,
  ``Bayesian-optimized bidirectional lstm regression model for non-intrusive
  load monitoring,'' in \emph{ICASSP 2019-2019 IEEE International Conference on
  Acoustics, Speech and Signal Processing (ICASSP)}.\hskip 1em plus 0.5em minus
  0.4em\relax IEEE, 2019, pp. 2747--2751.

\bibitem{feurer2019hyperparameter}
M.~Feurer and F.~Hutter, ``Hyperparameter optimization,'' in \emph{Automated
  machine learning}.\hskip 1em plus 0.5em minus 0.4em\relax Springer, Cham,
  2019, pp. 3--33.

\end{thebibliography}
%





\newpage
\onecolumn
\section*{Appendix}
\subsection{Notations}
  \label{FirstAppendix}
  \vspace{-0.4cm}
\begin{table*}[ht!]
\small
\renewcommand{\arraystretch}{1.2}
    \centering
    {\caption{Summarization of Notations}}
    \vspace{-0.1cm}
    \begin{tabular}{c|l|c|l}
         \hline\hline
         {\textbf{Notation}} & {\makecell[c]{\textbf{Description}}} & {\textbf{Notation}} & {\makecell[c]{\textbf{Description}}} \\
         \hline\hline
         {${cA,cH,cV,cD}$} & {Wavelet coefficient} & {${q_\sigma(\tilde{x}|x)}$ }& {Pre-specified noise distribution} \\
         \hline
         {${\nabla_xlogp(x)}$} & {Gradient of log density ${p(x)}$} & {${Y_t}$} & {The ${t}$-th Langevin diffusion} \\
         \hline
         {${S_\theta(\cdot)}$} & {Score-based network} & {${B_t}$} & {Standard Brownian motion} \\
         \hline
         {${p_{data}(x)}$ } & {Distribution of data ${x}$} & {${v_t}$} & {The law of ${Y_t}$}\\
         \hline
         {${y}$} &{Grayscale image} & {${z}$ }& {Random Gaussian noise} \\
         \hline
         {${x}$} & {Color image}& {${c_{LS}(\cdot)}$} &{A log-Sobolev inequality with constant}\\
         \hline
         {${x_R}$,${x_G}$,${x_B}$} & {${R}$, ${G}$, ${B}$ channel of ${x}$} & {${d^{'}}$} & {Intrinsic dimension}\\
         \hline
         {${\{\sigma_i\}^L_{i=1}}$} & {A series of noise level} & {${W(\cdot)}$} & {Discrete wavelet transform}\\
         \hline
         {${\lambda(\sigma_i)}$} & {Coefficient function depending on ${\sigma_i}$ } & { ${W_R,W_G,W_B}$ } & {Wavelet transform of ${x_R,x_G,x_B}$}\\
         \hline
         {${\alpha_i}$} & {Step size} & {${X}$ } & {Stack of ${W_R,W_G,W_B}$} \\
         \hline
         {${t}$} & {Number of iteration index} & {${\varepsilon}$} & {A parameter of step size} \\
         \hline
         {${T}$} & {Total number of iteration} & {${F(\cdot)}$} & {Degenerate function} \\
         \hline\hline
    \end{tabular}
    \label{notation}
\end{table*}

\subsection{Derivation of the Theorems}
  \label{secondAppendix}
  
\textbf{Theorem 1 }(Theorem 1 from \cite{block2020fast}). Let ${d\geq3}$ and suppose that the scores of ${p}$ and ${p_\sigma^2}$ are L-Lipschitz and ${(m,b)-dissipative}$. Let ${S_\theta}$ be an estimate of the score of ${p_{\sigma^2}}$ whose expected squared error with respect to ${p_{\sigma^2}}$ is bounded by ${\varepsilon^2}$. Suppose that ${X_0-\hat{v_0}}$. Under technical conditions on ${\hat{v_0}}$ satisfied by a multivariate Gaussian, we have 
\begin{equation}\begin{split}\label{appendix1}\begin{aligned}
    W_2(\hat{V_{t}},p) \leq (\sigma\sqrt{d}+W_2(\hat{V_0},p_{\sigma^{2}})e^{-\frac{2t}{c_{LS}(p_{\sigma^{2}})}}) +\\ C\sqrt{(b+d)}(\varepsilon t+ \Vert p_{\sigma^{2}} \Vert^{\frac{1}{2}-\frac{1}{d}}e^{\frac{L\sqrt{d}}{4}t}\sqrt{t}\varepsilon ^{{\frac{1}{d}}{\frac{1}{4}}}),
 \end{aligned}
 \end{split}
\end{equation}
where does not depend on the dimension. As can be seen, the bound of the Wasserstein distance in Eq. (\ref{appendix1}) is determined by the intrinsic subspace dimension. Furthermore, under Assumption 1, the bound will be simpler and more precise.

\textbf{Assumption 1.} Let ${M,g}$ be a ${d^{'}-dimensional}$, smooth, closed, complete, connected Riemannian manifold isometrically embedded in ${\mathbb{R}^d}$ and contained in a ball of radius ${\rho}$, such that there exists a ${K \leq 0}$ such that ${Ric_{M} \succcurlyeq -Kg}$ for all ${y \in M}$ in the sense of quadratic forms. With respect to the inherited metric, ${M}$ has a volume form vol, which has finite total integral on ${M}$ due to compactness. Then ${p=pvol_M}$ is continuous with respect to the volume form and we refer to its density with respect to this volume form as ${p}$ as well, by abuse of notation.

\textbf{Theorem 2 }(Theorem 3 from \cite{block2020fast}).  Suppose that the pair ${(M,g)}$  satisfies Assumption 1 and let ${p \varpropto vol_M}$ be uniform on ${M}$. Assume that ${K>1}$ and that ${\kappa>1}$, then
\begin{equation}\label{a2}\begin{aligned}
    {c_{LS}{(p_\sigma^{2})}=O{\left(\sigma^{2}+K^{4}{d^{'}}^{2}\kappa^{20K^{2}d^{'}}\right).}}
    \end{aligned}
\end{equation}

It should be emphasized that the above bound is completely intrinsic to the geometry of the data manifold and that the dimension of the feature space does not appear, thus we can conclude that even with arbitrarily high dimension in pixel space, if the feasible space has small dimension ${d^{'}}$, Langevin dynamics will still mix quickly.

\subsection{The Influence of Parameters on Colorization}
  \label{secondAppendix}
\begin{figure*}[ht!]
\setlength{\belowcaptionskip}{-3pt}
  \begin{center}
  \includegraphics[width=7in]{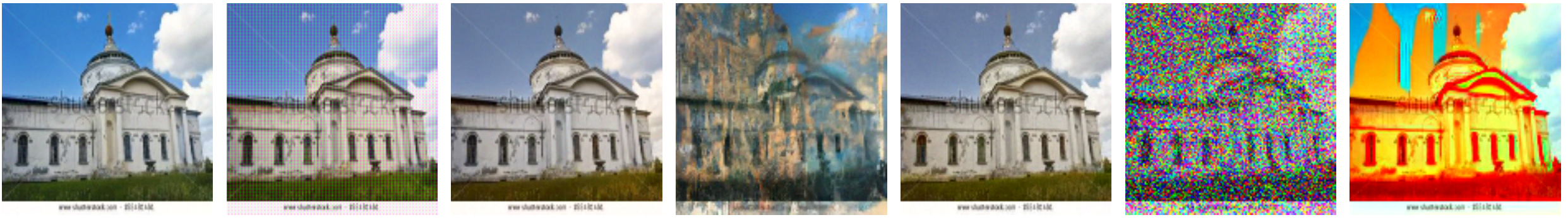}\\
  \vspace{-0.1cm}
  \leftline{ \hspace{0.3cm}(a) WACM \ \hspace{0.5cm} (b) 10${\times}$SC  \ (c) SC on low-frequency (d) 0.01${\times}$DC \ \hspace{0.2cm} (e) 12${\times}$DC \ \hspace{0.3cm} \ \ (f) 0.01${\times \varepsilon}$ \hspace{0.5cm} \ \ (g) 10${\times \varepsilon}$}
  \caption{\hspace{-0.6em}The influence of different parameters on colorization. Appropriate parameters of WACM contribute to the final high-quality colored results.}\label{app}
  \end{center}
\setlength{\belowcaptionskip}{-1pt}
\vspace{-0.8cm}
\end{figure*}


\vfill


\end{document}